\documentclass{article}

\usepackage{PRIMEarxiv}

\usepackage[utf8]{inputenc} 
\usepackage[T1]{fontenc}    
\usepackage{hyperref}       
\usepackage{url}            
\usepackage{booktabs}       
\usepackage{amsfonts}       
\usepackage{nicefrac}       
\usepackage{microtype}      
\usepackage{lipsum}
\usepackage{fancyhdr}       
\usepackage{graphicx}       
\graphicspath{{media/}}     
\newtheorem{definition}{Definition} 
\newtheorem{theorem}{Theorem}
\newtheorem{lemma}{Lemma}
\newtheorem{remark}{Remark}

\usepackage{amsmath}
\usepackage{amssymb}
\usepackage{multirow}
\usepackage{algorithm}
\usepackage{algpseudocode}
\usepackage{subcaption}
\usepackage{makecell}
\title{GFlowCausal: Generative Flow Networks for Causal Discovery
}

\author{
  Wenqian Li$^{1}$, Yinchuan Li$^{2}$\thanks{Corresponding author: Yinchuan Li}, Shengyu Zhu$^{2}$, Yunfeng Shao$^{2}$, Jianye Hao$^{2}$, Yan Pang$^{1}$   \\
  $^{1}$National University of Singapore \\
  $^{2}$Huawei Noah's Ark Lab\\
    \texttt{wenqian@u.nus.edu},\texttt{jamespang@nus.edu.sg}\\\texttt{\{liyinchuan,zhushengyu,shaoyunfeng,haojianye\}@huawei.com}\\ 
}

\begin{document}
\maketitle

\begin{abstract}
Causal discovery aims to uncover causal structure among a set of variables. Score-based approaches mainly focus on searching for the best Directed Acyclic Graph (DAG) based on a predefined score function. However, most of them are not applicable on a large scale due to the limited searchability. Inspired by the active learning in generative flow networks, we propose a novel approach to learning a DAG from observational data called GFlowCausal. It converts the graph search problem to a generation problem, in which direct edges are added gradually. GFlowCausal aims to learn the best policy to generate high-reward DAGs by sequential actions with probabilities proportional to predefined rewards. We propose a plug-and-play module based on transitive closure to ensure efficient sampling. Theoretical analysis shows that this module could guarantee acyclicity properties effectively and the consistency between final states and fully-connected graphs. We conduct extensive experiments on both synthetic and real datasets, and results validate the proposed approach to be superior and also performs well in a large-scale setting. 
\end{abstract}


\section{Introduction}
{U}{ncovering} causal relations from observational data is a burgeoning topic in machine learning and artificial intelligence, which has various applications such as biology~\cite{sachs2005causal} and genetics~\cite{peters2017elements}.  Some work formulates this causal discovery task as exploring a directed acyclic graphs (DAG) that could most represent the causal relationship from observational or experimental data~\cite{chickering1996learning,chickering2002optimal}. Even though some notable works are reluctant to treat graphs learned from observational data as causal~\cite{freedman1999there,dawid2010beware}, or consider ignoring the possibility of cycles will potentially introduce bias~\cite{strobl2019constraint}, it is still a common and worthwhile aim to begin with such well-known class. With certain assumptions, causal structure is fully or partially identifiable. For example, if one assumes the distribution is entailed by a structural equation model, then such graph is identifiable with mutual independence of noises and some mild regularity conditions~\cite{peters2014causal,lachapelle2019gradient}.

 

Depending on the way to solve, causal discovery can be categorized into four different classes: constraint-based~\cite{spirtes2000causation,chickering2002optimal}, score-based~\cite{zheng2018dags,zhu2019causal}, structural asymmetry-based~\cite{mooij2016distinguishing} and intervention-based~\cite{peters2017elements}. Most constraint-based approaches test for conditional independencies in the empirical joint distribution, while they require large sample sizes to be reliable. For example, PC ~\cite{spirtes2000causation} and Greedy Equivalence Search (GES) \cite{chickering2002optimal} rely on local heuristics for searching. Intervention-based approaches~\cite{peters2017elements} talk about that if interventional data are available, we are able to reduce the number of graphs in the markov equivalence class. Structural asymmetry-based tests the direction of edges between each pair of nodes~\cite{mooij2016distinguishing}. Score-based aims to compute a score on each graph over the observed data, and then search for DAGs with the best score.
However, searching for DAGs is known to be NP-hard~\cite{chickering1996learning} due to its combinatorial nature with acyclicity constraints and super-exponentially growing in the number of graphs. NOTEARS~\cite{zheng2018dags} firstly casts this problem into a continuous constrained one with the least-squares objective for linear models. Subsequently a flurry of developments in continuous optimization approaches includes DAG-GNN~\cite{yu2019dag}, NO FEARS~\cite{wei2020dags}, and GraN-DAG~\cite{lachapelle2019gradient}. Without violating some general assumptions, optimal ordering learning problems are adopted to guarantee the significantly smaller search space and avoid acyclicity constraints~\cite{teyssier2012ordering}. As another attempt, some research takes the strong searching ability of Reinforcement Learning (RL) for causal discovery. For example, RL-BIC \cite{zhu2019causal} utilizes predefined score functions and find the underlying DAG . Nevertheless, due to the large search space and inefficient acyclicity constraints, most searching approaches do not handle large-scale settings and get stuck on local optimum. {Furthermore, CORL~\cite{wang2021ordering} considers to take advantage of the reduced space of variable orderings, to make RL approach perform better than other gradient-based methods that directly optimize the same score function. Recently, Generative Flow Network based approach DAG-GFlowNet is proposed by~\cite{deleu2022bayesian}, which treats the generation of a sample as a sequential decision problem and use flow matching conditions to train a model, thus modeling the posterior distribution over DAGs. However, the trajectory length and loss objective matter in computational complexity and DAG-GFlowNet may not able to achieve a close performance than others in beyesian structure learning. }

In this paper, we consider DAG learning task in causal discovery, and take advantage of \textit{Generative Flow Networks} (GFlowNets) to cast the searching problem into the generation problem. Instead of directly searching DAGs and evaluating them, our idea is to gradually add a direct edge between each pair of two nodes, until a desired graph is generated. We call this approach  \texttt{GFlowCausal}, which could generate diverse DAGs by sequential actions with probabilities proportional to a predefined reward function. Our GFlowCausal could overcome the current predicament for the following reasons. First, it aims to learn the best policy to generate diverse high reward DAGs based on the flow matching condition, different from other searching or optimizing problems. This helps to avoid getting stuck in a sub-optimal situation. Second, with a strong exploration ability, GFlowCausal could be applied in a large scale setting and obtain convergence fast. Moreover, GFlowCausal defines allowed actions to ensure acyclicity in every state transition based on matrix calculations and reduce the computational complexity significantly to $\mathcal{O}(1)$ per operation. 
\subsection{Main Contributions} 
1) We introduces GFlowNets as another attempt to model causal discovery as a generation problem, rather than a searching or optimization problem; The proposed order graph with a much smaller search space could improve the efficiency of causal structure search. 2)~We give a formal definition of GFlowCausal, and provide a general architecture to solve the graph generation problem, where the types of neural networks and reward functions are not limited; 3) We propose an effective and efficient method to guarantee the acyclicity constraint in DAGs, which is a plug-and-play module and suitable for any DAG generation problems. We provide theoretical analysis to validate this, and also show the consistency between the final states and fully-connected graph;  4) We analyze the computational complexity of the proposed architecture and show how to reduce it further for different purposes; {5) We provide rigorous analysis on different generation cases, including both exact graph generation and order generation, and conduct different ablation experiments to intuitively explain the reasonable of our design.} 6) We conduct extensive experiments to show GFlowCausal can outperform current state-of-the-art methods.

The rest of our paper is organized as follows. In Section~\ref{relatedwork}, we introduce the related works about directed acyclic graph structure learning in causal discovery and generative flow networks, and also provide preliminary knowledge of GFlowNets for ease of understanding of our proposed algorithm. Section~\ref{gflowcausalsection} formulates the problem in formal and the general framework of GFlowCausal, including the state, action, reward design and network architecture. Section~\ref{Training Procedure and Computational Analysis} shows the training procedures and strategies to save the computational costs. Section~\ref{experiments1} provides the implementation details of GFlowCausal and the experimental results on different causal relationships. We finally conclude this work in Section~\ref{conclusion}.

\section{Related Work} 
\label{relatedwork}
\subsection{Causal Discovery}
In principle and under some assumptions, graphical frameworks could help us to answer scientific questions such as "how?","why?", and "what if?"~\cite{pearl2018book}. {One major branch of causal discovery aims to uncover an adjacency matrix representing the DAG of the joint distribution over the dataset, with an assumption that the underlying
that the underlying graph connecting the variables is a directed acyclic graph (DAG). Since this assumption could provides the solution to narrow down the class of graphs which are compatible with the observed probability distribution. This branch could also be divided into two different classes. One is the constraint-based approach, and the other one is the score-based approach.}  Constraint-based approaches construct graphs by testing conditional independencies in the empirical joint distribution, while they require large sample sizes to be reliable. Score-based approaches test the validity of candidate graphs according to a predefined score function. Some common score functions include Bayesian Information Criterion (BIC)~\cite{geiger1994learning}, Bayesian Gaussian equivalent (BGe)~\cite{geiger1994learning} and Bayesian Dirichlet equivalence (BDe)~\cite{heckerman1995learning} and Evidence Lower BOund (ELBO) \cite{kingma2013auto,yang2021causalvae},etc.

Traditional works such as~\cite{chickering2002optimal,lam1994learning} employ combinatoric approaches to identify a DAG structure. However, the search complexity is $O(2^{d^2})$, where $d$ is the number of nodes, which is an NP-hard problem. The first work to recast the combinatoric graph search problem as a continuous optimization problem is proposed by~\cite {zheng2018dags}, in which a smooth function quantifies the ``DAG-ness'' of the graph. Due to its high complexity and NP-hard characterization, further researches seek to improve on this or relax the constraint~\cite{yu2019dag,lachapelle2019gradient,kalainathan2018structural,moraffah2020causal,kyono2020castle,ng2020role}. Some work takes a reinforcement learning method to causal discovery, such as RL-BIC proposed by~\cite{zhu2019causal}. Authors generate directed graphs using an encoder-decoder neural network model as the ``actor'' and using BIC score and some penalty terms for cycles in graphs as a reward signal to score these candidate graphs. Further, \cite{wang2021ordering} extends this work to ordering-based reinforcement learning (CORL) to learn the optimal order of variables. In order to further reduce the problem size. Recently, \cite{yang2022reinforcement} proposes RCL-OG, which is a reinforcement learning based approach replacing the Markov chain Monte Carlo method with the order graph to model different DAG
topological orderings.

\subsection{Generative Flow Networks}
Generative flow networks are generative policies that could sample compositional objects $x \in \mathbb{D}$ by discrete action sequences with a probability proportional to a given reward function. This network could sample the distribution of trajectories with high rewards and can be useful in tasks when exploration is important. This is different from RL, which aims to maximize the expected return and only generates a single sequence of actions with the highest reward. GFlowNets has been applied in molecule generation~\cite{bengio2021flow}, discrete probabilistic modeling~\cite{zhang2022generative}, bayesian structure learning~\cite{deleu2022bayesian}, biological sequence design~\cite{jain2022biological} . There are some overlaps on modelling  between ours and the work in~\cite{deleu2022bayesian} and we show their differences in Appendix~\ref{concurrentwork}. Recently, ~\cite{malkin2022trajectory} proposes the trajectory balance loss as another training objective to accelerate the training process and convergence.

\subsection{Preliminary Knowledge of GFlowNets}  

For clarity, we present the background of GFlowNets here. For more details, we recommend readers to refer to \cite{bengio2021gflownet}. Considering a directed graph with tuple $(\mathcal{S},\mathcal{A})$, where $\mathcal{S}$ is a finite set of states, and $\mathcal{A}$ is a subset of $\mathcal{S} \times \mathcal{S}$ representing directed edges. In particular, $\mathcal{A}$  is named as the {\em \textbf{action set}} consisting edges or transitions $a: s_t \rightarrow s_{t+1}$. A {\em \textbf{trajectory}} in such a graph is defined as a sequence $(s_1,...,s_n)$ of elements of $\mathcal{S}$. In addition, given a directed acyclic graph (DAG), i.e., a directed graph in which there is no trajectory $(s_1,...,s_n)$ satisfying $s_n = s_1$ besides trajectories composed of one state only, a {\em \textbf{complete trajectory}} is defined as a sequence of states $\tau = (s_0,...,s_f) \in \mathcal{T}$ in which $s_0$ is the only {\em \textbf{initial state}}, $s_f$ is the {\em \textbf{final state}}, and $\mathcal{T}$ is the set of complete trajectories associated with such a given DAG.
Furthermore, the {\em \textbf{terminating state}} is defined as the penultimate state of the trajectory, e.g., $s_n$ is the terminating state of $(s_0, s_1,...,s_n,s_f)$.

\begin{definition}[Edge Flow \cite{bengio2021gflownet}]
An edge flow $F(s_t \rightarrow s_{t+1})$ is the flow through an edge $s_t \rightarrow s_{t+1}$.
\end{definition}

\begin{definition}[Terminating Flow \cite{bengio2021gflownet}]
A transition $s \rightarrow s_f$ into the final state is defined as the terminating transition, and $F(s\rightarrow s_f)$ is a terminating flow.
\end{definition}

\begin{definition}[Trajectory Flow \cite{bengio2021gflownet}]
A trajectory flow $F(\tau): \tau \mapsto \mathbb{R}^+$ is defined as any nonnegative function defined on the set of complete trajectories $\tau$. For each trajectory $\tau$, the associated flow $F(\tau)$ contains the number of particles sharing the same path $\tau$.
\end{definition}

\begin{definition}[State Flow \cite{bengio2021gflownet}]
The state flow $F(s): \mathcal{S} \mapsto \mathbb{R}$ is the sum of the flows of the complete trajectories passing through that state:
$$
F(s) = \sum_{\tau \in \mathcal{T}} 1_{s\in \tau} F(\tau).
$$
\end{definition}

\begin{definition}[Action Flow]
The action flow $F(s_t,a_t): \mathcal{S} \times \mathcal{A} \mapsto \mathbb{R}$ is the flow of the complete trajectory passing through state $s_t$ with a specific action $a_t$. When $a_t: s_t \rightarrow s_{t+1}$, $F(s_t,a_t)$ is equal to the edge flow $F(s_t \rightarrow s_{t+1})$.
\end{definition}

\begin{definition}[Inflows and Outflows]
For any state $s_{t+1}$, its inflows are the flows that pass through state $s_{t}$ and can reach state $s_{t+1}$, i.e., $\sum\nolimits_{s_t,a_t:T(s_t,a_t) = s_{t+1}} F(s_t,a_t)$ is the sum of inflows, where $T(s_t,a_t) = s_{t+1}$ indicates an action $a_t$ that could make a transition from state $s_t$ to attain $s_{t+1}$. In addition, the outflows are the flows passing through state $s_{t+1}$ with all possible actions $a_{t+1} \in \mathcal{A}$, i.e., $\sum\nolimits_{a_{t+1} \in \mathcal{A}}F (s_{t+1},a_{t+1})$ is the sum of outflows.
\end{definition}

\begin{definition}[Transition Probability \cite{bengio2021gflownet}]
The transition probability $\mathcal{P}(s_t \rightarrow s_{t+1}|s_t)$ is a special case
of conditional probability, which is defined as
$$
\mathcal{P}(s_{t+1}|s_t) := \mathcal{P}(s_t \rightarrow s_{t+1}|s_t) = \frac{F(s_t\rightarrow s_{t+1})}{F(s_t)}.
$$
\end{definition}



\begin{definition}[GFlowNet \cite{bengio2021gflownet}]
A GFlowNet is a pair $(\hat F(s); \hat{\mathcal{P}}(s_{t+1}|s_t) )$ where $\hat F(s)$ is a state flow function and $\hat{\mathcal{P}}(s_{t+1}|s_t)$ is a transition distribution from which one can draw trajectories by iteratively sampling each state given the previous one, starting at initial state $s_0$ and then with $s_{t+1} \sim \hat{\mathcal{P}}(s_{t+1}|s_t)$ for $t = 0, 1, ...$ until final state $s_{n+1} = s_f$ is reached for some $n$.
\end{definition}

\section{GFlowCausal}
\label{gflowcausalsection}
We develop problem formulation of causal structure learning in a step-by-step generative way in Section~\ref{problemformulation}. Then we specify the states and allowed actions in~\ref{stateandaction} and propose the transition dynamic property based on the transitive closure. After that, we provide two reward functions for evaluating whether the generated DAG fit the data well in Section~\ref{rewardfunction}.  
\begin{table}
\centering
\setlength{\abovecaptionskip}{0.05cm}
\setlength{\belowcaptionskip}{-0.3cm}
\setlength{\tabcolsep}{0.9 mm}{
\caption{Empirical results on LG data models with 100-node graphs }
\label{table_notation}
\begin{tabular}{ccc} 
\toprule
& $\textbf{Notations}$ &   $\textbf{Descriptions}$    \\ 
& $X$ & $n \times d$ data matrix  \\
& $x_i$ & Data vector consisting of $n$ i.i.d samples, $i=1,...,d$  \\
& $A$ & $d \times d $ binary adjacency matrix  \\
& $W$ & Weighted adjacency matrix related to $A$  \\
& $\mathcal{G}$&  Directed acyclic graph (DAG) \\
& $\mathbb{D}$&  Discrete space of DAGs \\
& $\mathcal{V}$&  Vertices in a DAG \\
& $\mathcal{E}$&  Edges in a DAG \\
& $v_{i\rightarrow j}$& A directed edge from node $i$ to node $j$ \\
& $\mathcal{Q}$ & Identifying matrix for stopping sampling  \\
& $H$ & $d \times d $ transitive closure matrix  \\
& $M$ & $d \times d $ masked matrix for allowed actions $\mathcal{A}$  \\
& $s_t$ & A state corresponding to the adjacency matrix $A(s_t)$ \\
& $\tau$ & A sequence of states $(s_0,s_1,...,s_f)$ \\
& $s_0$ & All zero-valued matrix  \\
& $s_f$ & Final state in a trajectory  \\
& $a_t$ & An action to add a directed edge $v_{i\rightarrow j}$ \\
& $\mathcal{V}^\sharp(v_i)$ &  The ancestor set of node $v_i$ \\
& $\mathcal{V}^\flat(v_i)$ &  The descendent set of node $v_i$ \\
& $r(\cdot)$ & reward function \\
& $F(\cdot)$ & Non-negative function denoting flows \\
& $\mathcal{P}(\cdot)$ & Probability measurement over the flow $F(\cdot)$\\
& $\pi(a_t|s_t)$ & trained policy, given state $s_t$ taking action $a_t$ \\
\bottomrule
\end{tabular}}
\end{table}
\subsection{Problem Formulation}
\label{problemformulation}

Direct acyclic relationship is crucial to causal structure learning. In real-world, many works consider use the smoothness function as the either hard or soft constraint for combinatorial optimization problem, thus having high computational costs. In addition, could not apply well in the large scale setting. Therefore, efficiency training could be a basic task. Diverse results could help to prevent from getting stuck in local optima has shown to be effective in~\cite{bengio2021flow}, but trajectory length matters to the efficiency, which would be a bottleneck if we direct use the GFlowNet structure. Motivated by this, we propose to combine topological sequence learning and GFlowNet to tackle above problems. Table~\ref{table_notation} summarizes the notations used in this paper.

Let $X \in \mathbb{R}^{n\times d}$ be a data matrix consisting of $n$ i.i.d samples of the random vectors $x_1,...,x_d$, where $x_i$ indicates the data vector of the node $v_i$. Let $\mathbb{D}$ denote the discrete space of DAGs $\mathcal{G}=(\mathcal{V},\mathcal{E})$, where $\mathcal{V} = \{v_i\}_{i=1}^d$ denotes the set of nodes and $\mathcal{E} = \{ v_{i\rightarrow j} \}_{i,j=1}^d$ denotes the set of direct edges, i.e., $v_{i\rightarrow j}$ denotes an edge from $v_i$ to $v_j$. We model $X$ via a structural equation model (SEM) with binary adjacency matrix $A\in \{0,1\}^{d\times d}$, and use $W \in \mathbb{R}^{d \times d}$ to represent the corresponding weighted adjacency matrix related to $A$ when such a weighted matrix is meaningful, e.g., with linear SEMs or other special forms of structural equations. We assume the probability model related with $\mathcal{G}$ as $p(X) = D_{i=1}^d p (x_i|\text{parent}(x_i))$, which is entailed by a SEM of the form
\begin{equation}
    v_i = f_i(\text{parent}(v_i)) + \epsilon_i, ~ i=1,...,d,
\end{equation}
where $\text{parent}(v_i)$ denotes the parents of $v_i$, i.e., the set of variables $v_j \in \mathcal{V}$ if there is a direct edge $v_{j \rightarrow i}$; $f_i$ is the function mapping from $\text{parent}(v_i)$ to $v_i$; and $\epsilon_i$'s denote jointly independent additive noise variables. We assume structural minimality, which implies nonzero coefficients in $W$ define the structure of the ground truth $\mathcal{G}$.

\begin{figure}[ht]
\setlength{\abovecaptionskip}{0.0 cm} 
\setlength{\belowcaptionskip}{-0.2cm} 
  \centering 
 \includegraphics[width=3.5 in]{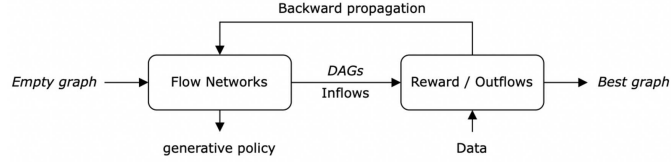}
  \caption{Structure of GFlowCausal.}
\label{figure1}
\end{figure}
Previous score-based approaches obtain graphs $\mathcal{G} \in \mathbb{D}$ directly from the dataset $X$, then search over the space to find the best graph $\mathcal{G}^{\star}$ according to a predefined score function $r(X,\mathcal{G})$, i.e.,
\begin{equation}
    \mathcal{G}^{\star} = \arg\max_{\mathcal{G} }\ r(X ,\mathcal{G}).
\end{equation}
As another attempt, we could turn this graph search problem into a step-by-step generative problem (see Figure~\ref{figure1}). The insight is that we could consider $\mathcal{G}$ as a compositional object. Starting from an empty graph, we can use a neural network as a sampler to generate such $\mathcal{G}$ by sequentially adding one direct edge between two nodes each time, which does not break the acyclicity constraint of a causal graph. After obtaining a fully-connected graph, we use a reward function $r(X ,\mathcal{G})$ to evaluate this graph. This process is similar to the episodic RL setting, in which each $\mathcal{G}$ refers to a state $s$, and adding one edge refers to an action $a$ making a state transition $s_t \rightarrow s_{t+1}$. The neural network is trained to learn the forward policy of such transitions. In particular, the dataset $X$ does not involve graph generation process but only to give a reward. Our output is a forward generative policy that advantages to sampling high-rewards DAGs.

We define this training structure as GFlowCausal, which consists of a tuple $(\mathcal{S},\mathcal{A})$ where $\mathcal{S}$ is a finite set of states, and $\mathcal{A}$ is the action set consisting transitions $a: s_{t} \rightarrow s_{t+1}$.  Let $F(\cdot)$ be a non-negative flow function, and $\mathcal{P}(\cdot)$ be the corresponding probability measurement over the flow, i.e., $\mathcal{P}(s_{t+1}\mid s_{t}) = \frac{F(s_t \rightarrow s_{t+1})}{F(s_t)}$ indicates the probability of a transition $ s_{t} \rightarrow s_{t+1}$.  A complete trajectory is defined as a sequence of states $\tau = {(s_0,s_1,...,s_f)} \in \mathcal{T}$ in which $s_0$ is the only initial state with $\mathcal{P}(s_0)=1$ and $s_f$ is the final state such that $\mathcal{G} = s_f$. Then, we have $\mathcal{P}(\tau) = \prod_{t=0}^{t=f-1} \mathcal{P}(s_{t+1} \mid s_t)$ and the flow passing through $s_f$ satisfies $F(s_f) = \sum_{\tau \in \mathcal{T}} \mathbb{I}_{s_f \in \tau} F(\tau)$ under the constraint that there is no cycles in the trajectory $\tau$, where $\mathbb{I}(\cdot)$ denotes the indicator function. Moreover, we could obtain
\begin{equation}
\mathcal{P}(s_f) = \frac{\sum_{\tau \in \mathcal{T}} \mathbb{I}_{s_f \in \tau} F(\tau)}{\sum_{\tau \in \mathcal{T}}F(\tau)} = \sum_{\tau \in \mathcal{T}}\mathbb{I}_{s_f \in \tau} \mathcal{P}(\tau).
\end{equation}
Suppose a policy $\pi : \mathcal{A} \times \mathcal{S} \mapsto \mathbb{R}$ is a probability distribution $\pi(a|s)$ over actions $a \in \mathcal{A}$ for each state $s$. Since in our setting only one particular direct edge could take the transition from $s_t$ to $s_{t+1}$, then $\pi(a_t|s_t) = \sum_{a:T(s_t,a)=s_{t+1}} \pi(a_t|s_t) = \mathcal{P}(s_{t+1}|s_t)$, where $T(s_t,a) = s_{t+1}$ indicates an action $a$ that could make a transition from state $s_t$ to attain $s_{t+1}$. Based on the above properties, we give the formal definition of GFlowCausal in Definition~\ref{def1}.

\begin{definition}[GFlowCausal]
\label{def1}
Given a dataset X, GFlowCausal aims to find the best forward generative policy $\pi(a_t \mid s_t)$ based on the flow network with parameter $\theta$ to generate $s_f$ by sequential actions with probabilities  
\begin{equation}
   \mathcal{P}_{\theta}(s_f) = \sum_{\tau \in \mathcal{T}}\mathbb{I}_{s_f \in \tau} \prod_{t=0}^{t=f-1}\mathcal{P}_{\theta}(s_{t+1} \mid s_t) 
= \sum_{\tau \in \mathcal{T}}\mathbb{I}_{s_f \in \tau} \prod_{t=0}^{t=f-1} \pi(a_t \mid s_t),
\end{equation}
which satisfies
\begin{equation}
\mathcal{P}_{\theta}(s_f) \propto r(s_f, X) 
\end{equation}
and there are no cycles in the trajectory $\tau$.
\end{definition}

With this nature, we could sample DAGs and evaluate whether they fit the observed dataset well. Therefore, it is crucial to specifically design the proper sampling strategies, including state and action, as well as the reward function in which better graph could obtain a higher reward. We will show our solutions in the following sections.

\subsection{States and Actions}
\label{stateandaction}
In this section we introduce details about the elements of GFlowCausal. 
Since GFlowCausal generates DAGs by sequentially adding direct edges, we can construct the binary adjacency matrix to model the DAGs and assign value 1 to represent adding one direct edge between two variables. With this intuition, first we give the following definitions on states and actions as follows,
\begin{definition}[State]
\label{state_def}
A state $s_t \in \mathcal{S}$ in GFlowCausal refers to an adjacency matrix $A(s_t)\in \{0,1\}^{d \times d}$. The initial state $s_0$ is an all-zero matrix corresponding to the empty graph and a final state $s_f$ is an adjacency matrix connecting to an identical fully-connected graph.
\end{definition}

\begin{definition}[Action]
\label{action_def}
An action (forward transition) $a: s_t \rightarrow s_{t+1} \in \mathcal{A}$ in GFlowCausal is to assign a position $A_{i,j}=1$ in $A(s_t)$, i.e. $T(A(s_t),A_{i,j}=1) = A(s_{t+1})$ with $A_{i,j}$ being the $(i,j)$-th element in $A$, which corresponds to add an edge $v_{j \rightarrow i}$ (denoted as $v_{j \rightarrow i}^{+}$) in the graph, under the constraint that there is no cycles in $\tau$. 
\end{definition}

By taking advantages of step-by-step generative process, we can define a constraint and only the allowed actions could make a transition between two states each time, to ensure that there is no cycle in the trajectory $\tau$. The intuition behind is that: starting from any node in a graphs, we can not pass through the same node in a sequence following the direct edges. Therefore, we should explore the ``ancestor-descendent'' relationships instead of only ``parent-child'' relationships among those nodes. Then based on this relationship, we construct a masked matrix to forbid adding value in positions that will break our acyclicity constraints. For any node $v_i$, we use ${\mathcal{V}^\sharp}(v_i)$ and $\mathcal{V}^\flat(v_i)$ to define its ancestor and descendent sets, respectively. Then, we propose a transitive closure structure to make it possible to answer ``ancestor-descendent'' relationships of nodes in $\mathcal{G}$. Based on Definition~\ref{state_def} and ~\ref{transitive_closure}, in the following Theorem~\ref{theorem 1} we show how to identify allowed actions. which is proved in the Appendix~\ref{Proof of the Proposition 1}. 



\begin{definition}
\label{transitive_closure}
A transitive closure (TC) of a state $s$ is a binary matrix $H(s) \in \{0,1\}^{d\times d}$, where $H_{i,j} = 1, i \neq j $ in $H(s)$ iff $v_i$ is reachable from $v_j$, i.e. $v_j \in  {\mathcal{V}^\sharp}(v_i)$ or  $v_i \in \mathcal{V}^\flat(v_j)$. 
\end{definition}

\begin{theorem}
\label{theorem 1}
Let $M(s_t) \in \{0,1\}^{d \times d}$ be a binary mask matrix, for any  action  $a_t = v_{i \rightarrow j}^{+} \in \mathcal{A}$, if
\begin{equation}
\label{mask_update}
    M(s_{t+1})= A(s_{t+1}) \vee H^{\mathsf{T}}(s_{t+1}),
\end{equation}
where $\vee$ denotes the disjunction operation, $(\cdot)^{\mathsf{T}}$ denotes the matrix transpose operator
and $H(s_0)$ is an identity matrix updated by
\begin{equation}
\label{TC_update_1}
    H(s_{t+1}) =  H(s_t) \vee h_{j}(s_t)\cdot h^{=}_{i}(s_t),
\end{equation}
with $h_{j}(s_t)$ being the $j$-th column vector of $H(s_t)$ corresponding to ${\mathcal{V}^\flat}(v_j)$ and  $h^{=}_{i}(s_t)$ being the $i$-th row vector of $H(s_t)$ corresponding to $\mathcal{V}^\sharp(v_i)$. Then, for $i,j = 1,...,d$, if $M_{i,j}(s_{t+1}) = 0$, we have $a_{t+1} = v_{j \rightarrow i}^{+} \in \mathcal{A}$ and if $M_{i,j}(s_{t+1}) = 1$, we have $a_{t+1} = v_{j \rightarrow i}^{+} \not\in \mathcal{A}$.  In this way we could further guarantee two acyclicity properties:\\
1)\ The GFlowCausal structure is a DAG, i.e. for $\tau = (s_0,s_1,...,s_f) \in \mathcal{T}$, $\forall s_t, s_{t+k} \in \tau$ with $k>0$, then $s_t \neq s_{t+k}$. And $\exists \tau_i, \tau_j \in \mathcal{T}, \tau_i \neq \tau_j$ such that $ s_f \in \tau_i, s_f \in \tau_j$;  \\
2)\ Every adjacency matrix $A(s_t)$ corresponds to a DAG $(\mathcal{V},\mathcal{E})$, i.e., $\forall v_i, v_j \in \mathcal{V}$, if $v_i \in \mathcal{V}^{\sharp}(v_j)$,  then $v_j \not\in \mathcal{V}^{\sharp}(v_i)$.
\end{theorem}

Theorem~\ref{theorem 1} shows how we utilize transitive closure to guarantee two acyclicity properties in both causal structure and GFlowCausal. The mask matrix identifies allowed actions to enforce acyclicity, if $M_{i,j}=0$, then $v_{j \rightarrow i}^{+} \in \mathcal{A}$ does not introduce any cycle. In contrast, if $M_{i,j} = 0 $, then $v_{j \rightarrow i}^{+} \not\in \mathcal{A}$.

\begin{remark}
To facilitate understanding, we give a simple example for the \eqref{TC_update_1}. Suppose $d=4$ and at state $s_1$ there exists an edge $v_{2 \rightarrow 3}$, thus $H_{3,2}(s_1)= 1$. Suppose $a_2 = v_{3\rightarrow 4}^{+} \in \mathcal{A}$ , then $H_{4,3}(s_2) = 1$. We also expect that $H(s_2)$ could identify the indirect relationship $v_2 \in \mathcal{V}^{\sharp}(v_4)$. 
At state $s_1$, we have $h_{4}(s_1) = [0\ 0\ 0\ 1]^{\mathsf{T}}$ and $ h^{=}_{3}(s_1) = [0\ 1\ 1\ 0]$, then we have 
$$
H(s_2) = \underbrace{\left[ 
\begin{array}{cccc}
1 & 0 & 0 & 0\\
0 & 1 & 0 & 0\\
0 & 1 & 1 & 0\\
0 & 0 & 0 & 1
\end{array}
\right]}_{H(s_1)}+ \underbrace{\left[ 
\begin{array}{cccc}
0 & 0 & 0 & 0\\
0 & 0 & 0 & 0\\
0 & 0 & 0 & 0\\
0 & 1 & 1 & 0
\end{array}
\right]}_{h_{4} \cdot\  h^{=}_{3}}  \\
=\left[ 
\begin{array}{cccc}
1 & 0 & 0 & 0\\
0 & 1 & 0 & 0\\
0 & 1 & 1 & 0\\
0 & 1 & 1 & 1
\end{array}
\right],
$$

in which we find $H_{4,2}(s_2) = 1$ indicating $v_2 \in \mathcal{V}^{\sharp}(v_4)$ and $H_{4,3}(s_2) = 1$ corresponding to $v_{3\rightarrow 4}$.
\end{remark}

\subsection{Reward Function}
\label{rewardfunction}
In Bayesian structure learning, there are various score functions to describe the fitness of the model. Given a predefined score function, GFlowCausal aims to train a sampler to sample high-rewards graphs with higher probabilities. Since the choice for the score function is flexible in our framework, we take BIC \cite{geiger1994learning} and varsortability\cite{reisach2021beware} as two examples below.

BIC score is one of the most common score functions, which is given by
\begin{equation}
\label{equation_bic_1}
S_{\text{BIC}}(s_f,X) = \sum_{i=1}^d\left[\sum_{k=1}^{n}\log p(x_i^k|\text{parent}(x_i^k);\theta_i)-\frac{|\theta_i|}{2}\log n\right]. 
\end{equation}
where $x_i^k$ is the $k-$th observation of $X_i$, and $\theta_i$ is the parameter associated with each likelihood. It scores a causal structure based on the likelihood of the data given that structure and penalizes with the degrees of freedom. The model with lower BIC score is better. To avoid the numerical issue of negative numbers, we use the following exponential term 
\begin{equation}
\label{equation_bic_2}
    r_{\text{BIC}}(s_f,X) = \exp(-S_{\text{BIC}}(s_f,X)).
\end{equation}
Alternatively, inspired by \cite{reisach2021beware}, with some certain assumptions, i.e. marginal variances could carry information about the causal order, we can optionally use {\textit{varsortability}} as another score function for our candidates. For any causal model over $d$ variables with a  adjacency matrix $A$, the varsortability is the fraction of directed paths that start from a node with strictly lower variance than the node they end in, that is,
\begin{align}
 &r_{\text{var}}(s_f,X) =\nu(\mathcal{G}) =  \frac{\sum_{k=1}^{d=1}\sum_{i \rightarrow j \in A^{k}} \gamma({\rm Var}(X_i),{\rm Var}(X_j))}{\sum_{k=1}^{d=1}\sum_{i \rightarrow j \in A^{k}} 1 } \nonumber\\
 &\gamma(a,b) =\left\{ 
\begin{array}{l}
1, ~~~~\ a<b\\
1/2, ~~ a=b\\
0, ~~~~\ a>b
\end{array} \right.,
\label{varsor_r1}
\end{align}
where $ \nu(\mathcal{G})\in (0,1) $. Varsortability equals one if the marginal variance of each node is strictly greater than that of its causal ancestors. We could multiply this term with a constant to make rewards more sparse. However, once the above variance condition breaks or the variance is unknown, the structure is not identifiable.

\begin{figure*}[!ht]
	\centering
	\includegraphics[width=6 in]{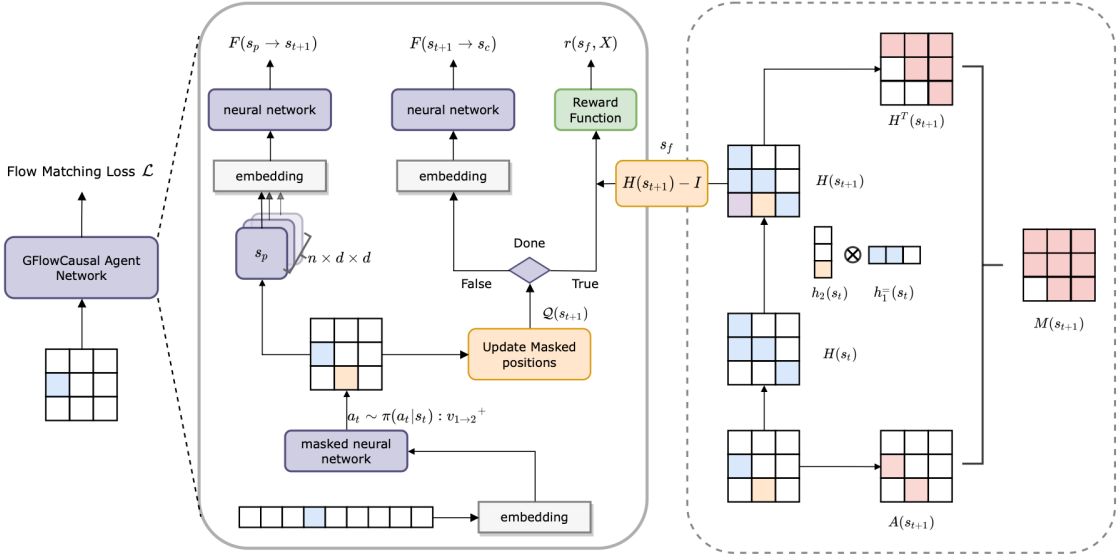}
 	\caption{Overall framework of GFlowCausal: The middle figure shows the sampling procedure and the right side figure shows the update of transitive closure matrix $H$, identifying matrix $\mathcal{Q}$ and masked matrix $M$. At each step $t$, the adjacency matrix at state $s_t$ is fed into the agent network to iteratively sample new DAGs until attain the stopping criteria. The defined neural network is trained to learn the $\mathcal{P}(a_t|s_t)$ with the flow matching loss $\mathcal{L}$. The embedding layers are optional for this task and the choices of neural networks are flexible depending on the task. }
	\label{networkstructure}
\end{figure*}

\section{Training Procedure and Computational Analysis}
\label{Training Procedure and Computational Analysis}

With increasing size of variables, the computational cost is the main challenge for causal discovery.  
In the following, we propose the GFlowCausal training framework and crucial theorems to address this problem, which includes the overall sampling procedure in Section~\ref{overallframe},the stopping criteria for efficiency sampling by specifying the corresponding relationship between the topological sorting and fully-connected graphs in Section~\ref{trainingprocedure}. In addition, the training objective will also be introduced to sample diverse candidates. Then we will provide different sampling strategies for different purposes and conduct computational analysis in Section~\ref{computationanalysis}.

\subsection{Overall Framework}
\label{overallframe}

The overview framework of GFlowCausal is shown in Figure~\ref{networkstructure}.
During the sampling phase (Middle part of Figure~\ref{networkstructure}), we sample an action probability buffer based on the forward-propagation of GFlowCausal, the neural network is masked since only valid actions could be sampled. After obtaining the action, the agent takes the action to update the state, and this process repeats until the complete trajectory is sampled, which corresponds to attaining the stopping criteria of the sampling procedure. For the inflows, we explore all pairs of parent states and corresponding one-step actions, and feed them into the neural network to approximate the total inflows. For the outflows or rewards, which depends on whether the current state is the last state of the complete trajectory, we feed the current state into the neural network to approximate the total outflows or use the predefined reward function to calculate the reward of this graph. 
Simultaneously, to guarantee the acyclicity of both trajectories and the causal structure, the masked matrix should be  updated based on the current state to provide valid actions (Right part of Figure~\ref{networkstructure}). The masked matrix consists of two parts: transitive closure and current state matrix, in which the transitive closure is constructed based on the current action and state. 
Finally, based on the approximated inflows and outflows, the flow matching loss $\mathcal{L}$ and its backpropagation are calculated for training the neural network. We name this process the flow matching procedure. The details about related theorem, stopping criteria and the training objective formula are shown in Section~\ref{trainingprocedure}.

\subsection{Training Procedure}
\label{trainingprocedure}
Starting from an empty graph, GFlowCausal draws complete trajectories $\tau = (s_0,s_1,...,s_f) \in \mathcal{T} $ by iteratively sampling $v_{i \rightarrow j}^{+} \sim \pi(a_t \mid s_t)$, until a fully-connected graph is generated. To simplify the sampling procedure, we consider this process as constructing the topological sort (see Definition~\ref{TS_def}) of the graph. We say the topological sort is {\em identified} once we obtain a complete topological sort of the graph, which could directly generate a fully-connected graph, thus we introduce an identifying matrix (see Theorem~\ref{theorem_2}) to check whether the topological sort is identified.  

\begin{definition} 
\label{TS_def}
For any state $s_t$, its corresponding topological sort ${\vec{\textbf{v}}}(s_t) = ( v_{k_1},...,v_{k_m} )$ with $m\leq d$ is a linear sequence containing partial elements in $\mathcal{V}$ such that $\forall a<b$, then $v_{k_a} \in \mathcal{V}^\sharp(v_{k_{b}})$. And a complete topological sort is defined as ${\vec{\textbf{v}}_{\star}} = (v_{k_1},...,v_{k_d} )$ containing all elements in $\mathcal{V} = \{v_i\}_{i=1}^d$.
\end{definition}

\begin{theorem}
\label{theorem_2}
Define $Q(s) \in \mathbb{R}^{d \times d}$ as an identifying matrix, which is updated by
\begin{equation}
\label{eq_11}
    Q(s_t) = H(s_t) \vee H^{\mathsf{T}}(s_t),
\end{equation}
where $H(s_t)$ is updated by \eqref{TC_update_1}. We have \\
1) $\vec{\textbf{v}}(s_t)$ is identified at $s_t$, i.e. $\vec{\textbf{v}}(s_t) = {\vec{\textbf{v}}_{\star}}$, iff $\forall i,j= 1 ,...,d, ~ Q_{i,j}(s_t) \neq 0$; \\
2) once $\vec{\textbf{v}}(s_t)$ is identified, then $\forall k >0$, $\vec{\textbf{v}}(s_{t+k}) = \vec{\textbf{v}}(s_t) = {\vec{\textbf{v}}_{\star}}$; \\
3) once $\vec{\textbf{v}}(s_t)$ is identified, then $ (H(s_t)-I_{d \times d})$ leads to a fully-connected graph.
\end{theorem}

Theorem 2 shows that we can judge whether the complete topological sort is identified or not based on \eqref{eq_11}, and once it is identified, continuing sampling will no longer update the topological sort. This indicates that the sampling of the current trajectory can be stopped.

After sampling a buffer, to train the policy $\pi(s_t\mid a_t)$ which satisfies $\mathcal{P}_{\theta}(s_f) \propto r(s_f,X)$, we minimize the loss over the flow matching condition
\begin{equation}
\begin{split}
\label{loss}
&{\mathcal{L}}_{\theta} (\tau)
=\sum\limits_{s_{t+1} \in \tau} \Bigg(  \sum\limits_{s_t,a_t:T(s_t,a_t) = s_{t+1}} F_{\theta} (s_t,a_t)  - \\
&\mathbb{I}_{s_{t+1} =  s_f}r(s_{t+1}, X)
- \mathbb{I}_{s_{t+1} \neq s_f} \sum\limits_{a_{t+1} \in \mathcal{A}}F_{\theta} (s_{t+1},a_{t+1}) \Bigg)^{2}.
\end{split}
\end{equation}
where $  \sum\nolimits_{s_t,a_t:T(s_t,a_t) = s_{t+1}} F_{\theta} (s_t,a_t) $ denotes the inflows of a state $s_{t+1}$, $\sum\nolimits_{a_{t+1} \in \mathcal{A}}F_{\theta} (s_{t+1},a_{t+1})$ denotes the outflows of $s_{t+1}$, and $r(s_f, X)$ denotes the reward of the final state, which can be \eqref{equation_bic_2} or \eqref{varsor_r1}.  For interior states, we only calculate outflows based on the neural network. For final states, there are no outgoing flows and we only calculate their rewards.

\begin{remark}
1) We find the process of evaluating graphs is dominating in the total running time. Compared to ~\cite{deleu2022bayesian}, our ordering generation approach does not spend much time computing rewards; 
2) If we evaluate DAG every state transition, which is the exact graph generation, we need to replace $\mathbb{I}_{s_{t+1} =  s_f}r(s_{t+1}, X)$ with $r(s_{t+1},X)$ in \eqref{loss}. However, the search space increases exponentially and it becomes more expensive to compute the loss function with this approach. 
\end{remark}

Compared to the ordering generation inspired by Theorem~\ref{theorem_2}, this exact graph generation approach has an exponential increasing size of search space, which is undesirable for exploration.

For simplicity, we summarize our GFlowCausal algorithm in Algorithm 1. Starting from an empty graph, for every iteration, GFlowCausal samples a valid action $a_t = v_{i\rightarrow j}^{+} \sim \pi(a_t \mid s_t)$ s.t. $M_{j,i}(s_t)=0$ based on the generative flow network $\theta$, and adds the direct edge to make a state transition $s_t \rightarrow s_{t+1}$.  Note that the final state $s_f$ corresponds to a fully-connected graph, a common practice is to prune it to get the final best graph. For linear data models, we can use thresholding to prune edges with small weights, as similarly used by \cite{zheng2018dags}, or prune edges based on the LASSO regression \cite{reisach2021beware}. For the non-linear model, we adopt the CAM pruning method \cite{lachapelle2019gradient}. For each variable $v_j$, one can ﬁt a generalized additive model against the current parents of $v_j$ and then apply significance testing of covariates, declaring significance if the reported $p$-values are lower then or equal to 0.001.

We can see the steps of transitions determine the trajectory length and thus the computational cost grows linearly. In the next subsection, we will show the computational analysis for this training procedure and explores to decrease the trajectory length with 3 different cases to show the efficiency of our proposed approach.

\subsection{Computational Analysis }
\label{computationanalysis}
\begin{lemma}
\label{lemma_1}
For any graph with $\{ \mathcal{V} \}_{i=1}^d$, at least $(d-1)$ transitions, at most $\frac{d(d-1)}{2}$ transitions could identify $\vec{\textbf{v}}_{\star}$, i.e. $\forall \vec{\textbf{v}}(s_t) = {\vec{\textbf{v}}_{\star}}$, then $(d-1) \leq t \leq d(d-1)/2$.
\end{lemma}

\begin{figure}
\centering
    \includegraphics[width=0.35\textwidth]{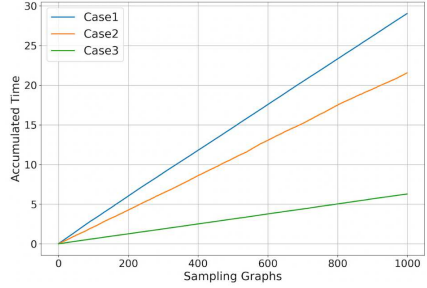}
    \caption{Time comparison with different cases. Our GFlowCausal is based on Case 2 and Case 3, which show the efficiency than that without Theorem~\ref{theorem_2}.}
\label{time_comparison}
\end{figure}

GFlowCausal could efficiently check valid actions with complexity $\mathcal{O}(1)$ for one hop according to \eqref{mask_update} and \eqref{TC_update_1}. Since for a complete trajectory leading $s_f$, we have $\frac{d(d-1)}{2}$ transitions, then the overall complexity for a trajectory is $\mathcal{O}(d^2/2)$. To reduce complexity, in the sampling procedure we transfer the structure generation problem to the topological identification problem by introducing \eqref{eq_11}, thus potentially smaller the length of the trajectory and decease the complexity to $\mathcal{O}(d^2/4)$ based on Lemma~\ref{lemma_1}, which is proved in Appendices~\ref{Proof of the Lemma 1} and \ref{Proof of the Lower Bound}. Furthermore, with some certain constraints, we could guarantee the complexity is $\mathcal{O}(d)$. For example, for first $(d-1)$ hops, in every transition we make sure a new node is connected and expand the topological sort. In Figure~\ref{time_comparison} we utilize three different sampling cases to sample 1000 graphs, and record the accumulated time. Case 1 does not introduce matrix $Q$ and each trajectory samples $\frac{d(d-1)}{2}$ transitions. Case 2 corresponds to \eqref{eq_11}. Case 3 has additional position constraints to guarantee $\mathcal{O}(d)$. Case 2 and Case 3 are applied to different proposes. The former evaluates different DAGs with the same topological sort, thus better identify the most important edges in causal structure. The later only evaluates fully-connected graphs and thus it is more suitable for ordering generation.

\begin{algorithm}[ht]
\small
\caption{GFlowNets for Causal Discovery} %
  \label{alg::conjugateGradient}
  \begin{algorithmic}[1]
    \Require
      Observed data $X$;
      $B$: batch size;
      $E$: epoch number;
      $\eta$: learning rate
    \State Initial $M(s_0) = I_{ d\times d}$, $H(s_0) = I_{ d\times d}$, $Q(s_0) = I_{ d\times d}$, $A(s_0) = \textbf{0}_{d \times d}$, and $\text{BestScore}=0$

    \Repeat
        \Repeat {~\emph{(parallel do with a batch size $B$)}}
        \State Sample a valid action $a:  v_{i\rightarrow j}^{+} \sim \pi(a_t \mid s_t)$ s.t. $M_{j,i}(s_t)=0$ based on the flow network ${\theta}$
        \State Make a state transition $s_{t+1} = T(s_t,a)$ and obtain $A(s_{t+1})$
        \State Update $H(s_{t+1})$, $M(s_{t+1})$ and $Q(s_{t+1})$ according to \eqref{TC_update_1}, \eqref{mask_update} and \eqref{eq_11}
        \Until{$Q$ is an all non-zero matrix}
        \State Calculate $r(s_f, X)$
            \If {$r(s_f,X)>$ BestScore}
            \State Update BestScore $\leftarrow r(s_f,X)$ and obtain a pruned graph $\mathcal{G}^{\star}$ based on $s_f$
        \EndIf
    \State Update  the network parameter ${\theta}$ based on $\nabla {\mathcal{L}}_{\theta} (\tau)$ and $\eta$ 
  \Until{epoch number $E$ is reached}
  \Ensure
        Policy $\pi(a_t \mid s_t)$ and the best graph $\mathcal{G}^{\star}$
  \end{algorithmic}
\end{algorithm}

\section {Experiments}
\label{experiments1}

In this section we will show the details about our implementation setting and experimental results. We conduct extensive experiments on both synthetic datasets in section~\ref{linearwithnnoise} and section~\ref{nolinearmodels}, as well as the real dataset in section~\ref{Experimental Results on Real Dataset}. Both linear and nonlinear models with gaussian and non-gaussian noises are included for synthetic datasets. The true DAGs are known to be identifiable for synthetic datasets, and synthetic datasets are generated based on these DAGs. The baselines are ICA-LiNGAM\cite{shimizu2006linear}, NOTEARS\cite{zheng2018dags}, DAG-GNN\cite{yu2019dag}, GraN-DAG\cite{lachapelle2019gradient}, GES\cite{chickering2002optimal} and two reinforcement learning approaches RL-BIC\cite{zhu2019causal} and CORL\cite{wang2021ordering} (see Appendix~\ref{Baselines} for details). We run experiments on three graph sizes $d \in \{12,30,100\}$ with  $\beta \in \{2,5\}$. Erd$\ddot{o}$s-R$\acute{e}$nyi (ER) and Scale-free (SF) are used to sample data from ground truth. $d$-node ER and SF graphs with on average $\beta \times d$ edges as ER$\beta$ and SF$\beta$. Three metrics are considered for evaluating a candidate graph: True Positive Rate (TPR), False Discovery Rate (FDR), and Structural Hamming Distance (SHD). SHD counts the total number of missing, falsely detected, or reversed edges, and lower indicates a better causal graph.

\subsection{Baselines for Comparisons} 
We introduce details about baselines in this section. The implmentation and codes are available in Appendix~\ref{Baselines}.

NOTEARS \cite{zheng2018dags} firstly attempts the continuous optimization to learn the underlying causal structure from data. it proposes the smooth characterization for acyclicity constraints, and recovers the causal graph via estimating the weighted adjacency matrix with the least squares loss with a predefined threshoding. 

ICA-LiNGAM \cite{shimizu2006linear} assumes linear Non-Gaussian additive model for data generating procedure and applies independent component analysis to recover the weighted adjacency matrix. 

S \cite{shimizu2006linear} recovers the causal graph by estimating the weighted adjacency matrix with the least squares loss and the smooth characterization for acyclicity constraint, followed by thresholding on the estimated weights.  

DAG-GNN \cite{yu2019dag} formulates causal discovery in the framework of variational autoencoder for nonlinear data. It uses a modified smooth characterization for acyclicity and optimizes a weighted adjacency matrix with the evidence lower bound as loss function.  

PC \cite{kalisch2007estimating} is a classic causal discovery algorithm based on conditional independence tests.  

GraN-DAG \cite{lachapelle2019gradient} models the conditional distribution of each variable given its parents with feed-forward NNs. It also uses the smooth acyclicity constraint from NOTEARS.  

RL-BIC2 \cite{zhu2019causal} proposes to use reinforcement learning to search for the DAG with the best score.  It considers encoder-decoder models for graph generation and incorporates both bayesian information criteira and the smooth acyclicity constraint from NOTEARS as the penalty for calculating the score of each DAG. However, the acyclicity constraint has $\mathcal{O}(d^3)$ complexity for each graph and it could not guarantee the valid graphs. The further penalties and related parameters make RL-BIC2 need carefully fine tuning.  

CORL\cite{wang2021ordering} formulates the causal discovery as causal ordering search problem. It establishes a canonical correspondence between the ordering and the fully-connected DAG, and then each step the RL solver selects one variable to model the markov decision process. After finding the ordering who could achieve the best score, it conducts the variable selections with pruning based on the significance testing.  

DAG-GFlowNet \cite{deleu2022bayesian} formulates the bayesian structure learning with GFlowNets, in which the edges are added gradually to construct direct acyclic graphs. The forward transition probability is defined as follows,
\begin{equation}
    \mathcal{P}_\theta(\mathcal{G}^\prime |\mathcal{G}) = (1-\mathcal{P}_\theta(s_f | \mathcal{G})) \mathcal{P}_\theta(\mathcal{G}^\prime | \mathcal{G},\lnot s_f),
\end{equation}
in which it distinguishes the terminal state and interior state.
This work proposes an adjusted flow matching objective, the detailed balance condition, to fit the stopping criteria of sampling as follows,
\begin{equation}
    \sum_{s\in Pa(s^\prime)} F_\theta(s\rightarrow s^\prime) - \sum_{s^{\prime\prime} \in Ch(s^\prime)} F_\theta(s^\prime \rightarrow s^{\prime\prime}) = R(s^\prime).
\end{equation}
When sampling invalid actions, for example, introducing cycle among variables or take a repeated action. For DAG-GFlowNet, every DAG is potential candidate and it chooses the DAG with high scores.

\subsection{Linear Models with Noise }
\label{linearwithnnoise}

For linear models, the synthetic datasets are generated in the following manner similar to that used by \cite{zhu2019causal,yu2019dag}. We first generate a random DAG by assigning weights $\text{Unif}([-2,-0.5] \cup [0.5,2])$ for the edges to obtain the weighted adjacency matrix $W \in \mathbb{R}^{d\times d}$. A sample dataset $X$ with $n=1000$ is generated by $X = W^{T}$X$ + Z$ from both Gaussian and non-Gaussian noise models. The true causal graph is generated by transforming $W$ into a binary matrix $A$.


First we consider the case of the Gaussian noise for ER2 with 12-nodes and 30-nodes. We use $r_{\text{var}}(s_f,X)$ defined in \eqref{varsor_r1} as the reward function. We set 5000 and 10000 epochs for 12-node and 30-node to train  2-MLP. The optimizer is Adam, and the learning rate is 0.0001. Our goal is to learn a generative policy to generate a diverse set of DAGs with high rewards. We do not care as much about the maximizing objective as in RL methods. After that, we use the trained policy to generate new 5000 graphs and count high rewards candidates. To compare GFlowCausal with RL-based ordering search method, we choose CORL with the same parameters. We select the MLP-encoder and LSTM-decoder for CORL architecture. 

\begin{figure*}[ht]
\setlength{\abovecaptionskip}{0.0cm} 
\setlength{\belowcaptionskip}{-0.0cm} 
\centering
\subcaptionbox{\label{Figure2a}}
{
\includegraphics[width=.29\linewidth]{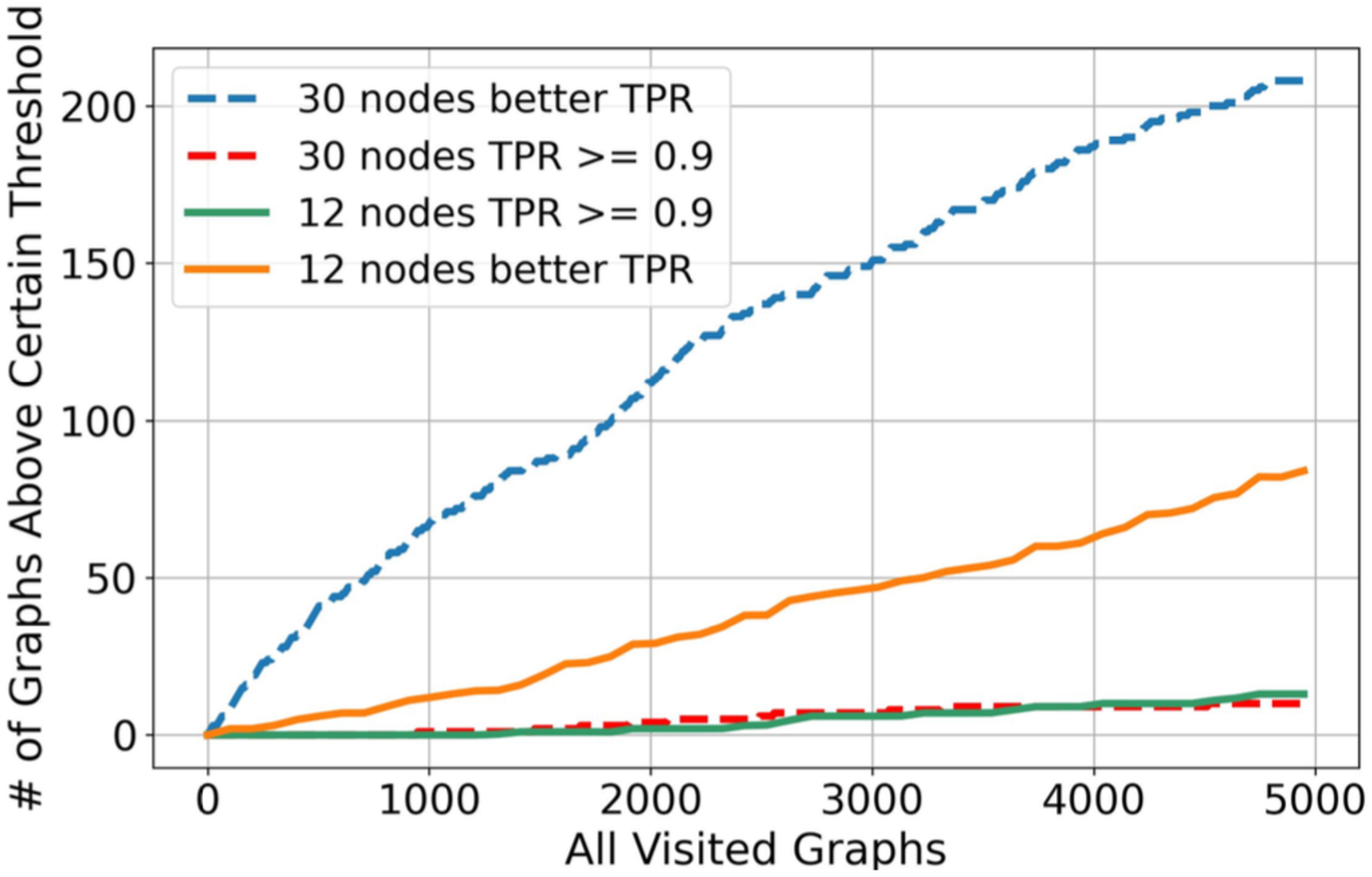}} 
\subcaptionbox{\label{Figure2b}}
{
\includegraphics[width=.32\linewidth]{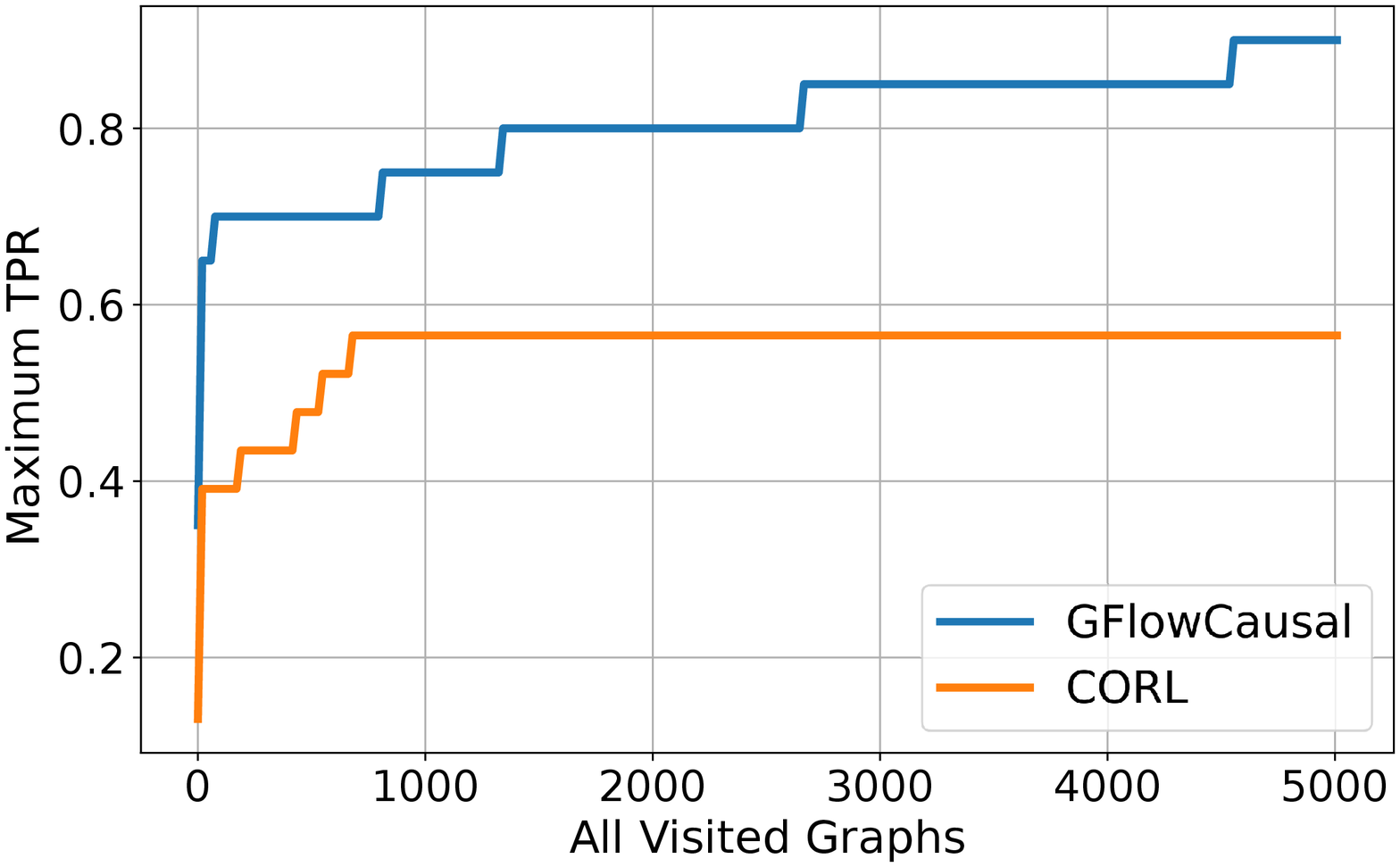}
}
\subcaptionbox{\label{Figure2c}}
{
\includegraphics[width=.29\linewidth]{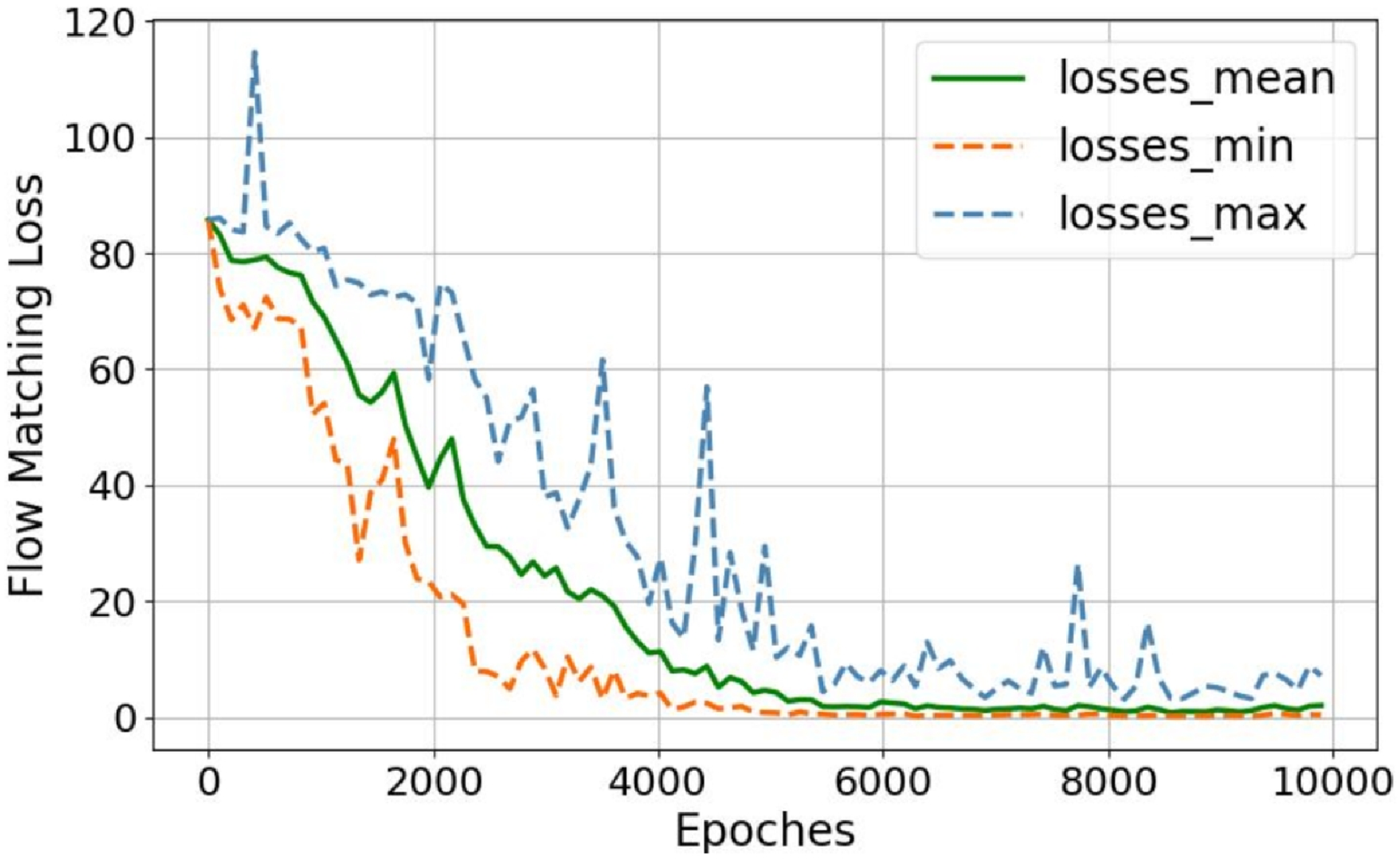}
}
\caption{(a) Number of distinct DAGs found above the certain threshold. (b) Maximum TPR graph sampled by GFlowCausal and CORL. (c) Training Loss of GFlowCausal on 100 nodes.  \label{comparison_CORL_GFLOW}}

\end{figure*}

\begin{table*}\small
\renewcommand\arraystretch{1.8}
\centering
\setlength{\abovecaptionskip}{0.cm}
\setlength{\belowcaptionskip}{-0.3cm}
\setlength{\tabcolsep}{0.5 mm}{
\caption{Empirical results on LG data models with 30-node graphs }
\label{table_1}
\begin{tabular}{cccccccccccc} 
\toprule
&  & ICA-LiNGAM   &  NOTEARS &   DAG-GNN    & GraN-DAG         & PC  &   RL-BIC2 & CORL           & Ours        \\ \hline
\multirow{2}{*}{ER2} &TPR&  0.49$\pm$0.07 &  0.93$\pm$0.03 &  0.92$\pm$0.08  & 0.13$\pm$0.06    &  0.43$\pm$0.12   &  0.92$\pm$0.05  & 0.92$\pm$0.04  &  \textbf{0.93$\pm$0.01}  \\
&SHD&  59.4$\pm$20   &  14.2$\pm$9.4 &   6.2$\pm$2.3  & 62.5$\pm$2.5    &  53.2$\pm$11.8  &   27.2$\pm$8.4    & 2.4$\pm$0.4    & \textbf{ 1.8$\pm$0.3} \\\hline
\multirow{2}{*}{ER5} &TPR& 0.69$\pm$0.08&  0.91$\pm$0.02 & 0.82$\pm$0.03       &  0.18$\pm$0.03   &  0.11$\pm$0.05  &      0.82$\pm$0.07       & 0.89$\pm$0.02  &   \textbf{0.92$\pm$0.06} \\
&SHD& 116.2$\pm$37.2 & \textbf{35.4$\pm$7.3} & 71.2$\pm$8.4        &  143.2$\pm$13.8     & 159$\pm$20.4  &   58.7$\pm$11.9   & 41.3$\pm$5.4    &  44.5$\pm$9.5 \\\hline

\multirow{2}{*}{SF2} &TPR&  0.71$\pm$0.06  &  0.91$\pm$0.07   &  0.92$\pm$0.04  & 0.30$\pm$0.03&   0.54$\pm$0.16  &  0.91$\pm$0.04   &  0.92$\pm$0.05  & \textbf{0.93$\pm$0.04}   \\
&SHD&  35$\pm$15   &  5.2$\pm$0.3  &  69.4$\pm$14.2   &  43.4$\pm$0.8  & 46.7$\pm$3.8     &  7.4$\pm$2.3   & 3.5$\pm$1.2 &  \textbf{2.1$\pm$0.4} \\\hline
\multirow{2}{*}{SF5} &TPR& 0.67$\pm$0.10 & 0.86$\pm$0.04   & 0.84$\pm$0.07 &  0.06$\pm$0.01    &   0.23$\pm$0.12      &  0.91$\pm$0.04&  0.91$\pm$0.02   &  \textbf{0.93$\pm$0.04}  \\
&SHD& 114.6$\pm$30.4  &  37.2$\pm$8.4  & 35.8$\pm$11.7  &    124.4$\pm$28.6   &    130.2$\pm$18.6    &  47.2$\pm$5.8 &  34.1$\pm$1.7   &  \textbf{28.4$\pm$3.6}   \\
\bottomrule
\end{tabular}}
\end{table*}


\begin{table*}[!ht]\small
\centering
\renewcommand\arraystretch{1.4} 
\setlength{\abovecaptionskip}{0.1cm}
\setlength{\belowcaptionskip}{-0.2cm}
\setlength{\tabcolsep}{0.6 mm}{
\caption{Empirical results on LG data models with 100-node graphs }
\label{table_2}
\begin{tabular}{cccccccccccc} 
\hline & &   ICA-LiNGAM   &  NOTEARS &  DAG-GNN & GraN-DAG  & PC &   RL-BIC2 & CORL  & Ours        \\ \hline
\multirow{2}{*}{ER5}&TPR   &   0.58$\pm$0.02    &  0.75$\pm$0.08       &     0.72$\pm$0.11     &  0.01$\pm$0.00 & 0.08$\pm$0.02 &   0.09$\pm$0.03      &   0.91$\pm$0.02  &   \textbf{0.95$\pm$0.01} \\ 
&SHD   &   599$\pm$18     &   281$\pm$87      &     207$\pm$43  &    514$\pm$7 &   595$\pm$19  &   434$\pm$50     &    209$\pm$22    &   \textbf{192$\pm$18}  \\  
\multirow{2}{*}{SF5}&TPR&  0.64$\pm$0.03  &  0.83$\pm$0.16  &  0.81$\pm$0.14  &  0.06$\pm$0.02 & 0.11$\pm$0.04 &   0.12$\pm$0.02  & 0.94$\pm$0.04  & \textbf{0.96$\pm$0.02}  \\

&SHD&  707$\pm$43    &  200$\pm$41   &  172$\pm$47   & 557$\pm$34   &  547$\pm$26  &   411$\pm$64 &  \textbf{47$\pm$17} & 58$\pm$19 \\
\bottomrule
\end{tabular}}
\end{table*}

Let us first look at what is learned by GFlowCausal. Figure~\ref{Figure2a} shows the number of graphs with high TPRs sampled by GFlowCausal. The dot lines show the results of 30 nodes, and the straight lines show that of 12 nodes. Both cases count the accumulated graphs with better TPR than the best graph generated during the training process and accumulated graphs with TPR greater than 0.9, which almost access the ground truth. As expected, GFlowCausal could generate more diverse candidates than CORL in Figure~\ref{Figure2b}. This exciting result gives us an insight about taking high-reward graphs into further training and thus the sampler knows better how to generate ``good enough'' candidates. This is important to a large-scale setting since it is unpractical to explore the entire huge space $\mathbb{D}$.
We also plot the training loss on 100 nodes in Figure~\ref{Figure2c} with batch size 128, in which we obtain the convergence of about 7000 epochs.
\begin{figure}[ht]
	\centering
	\includegraphics[width=2.5 in]{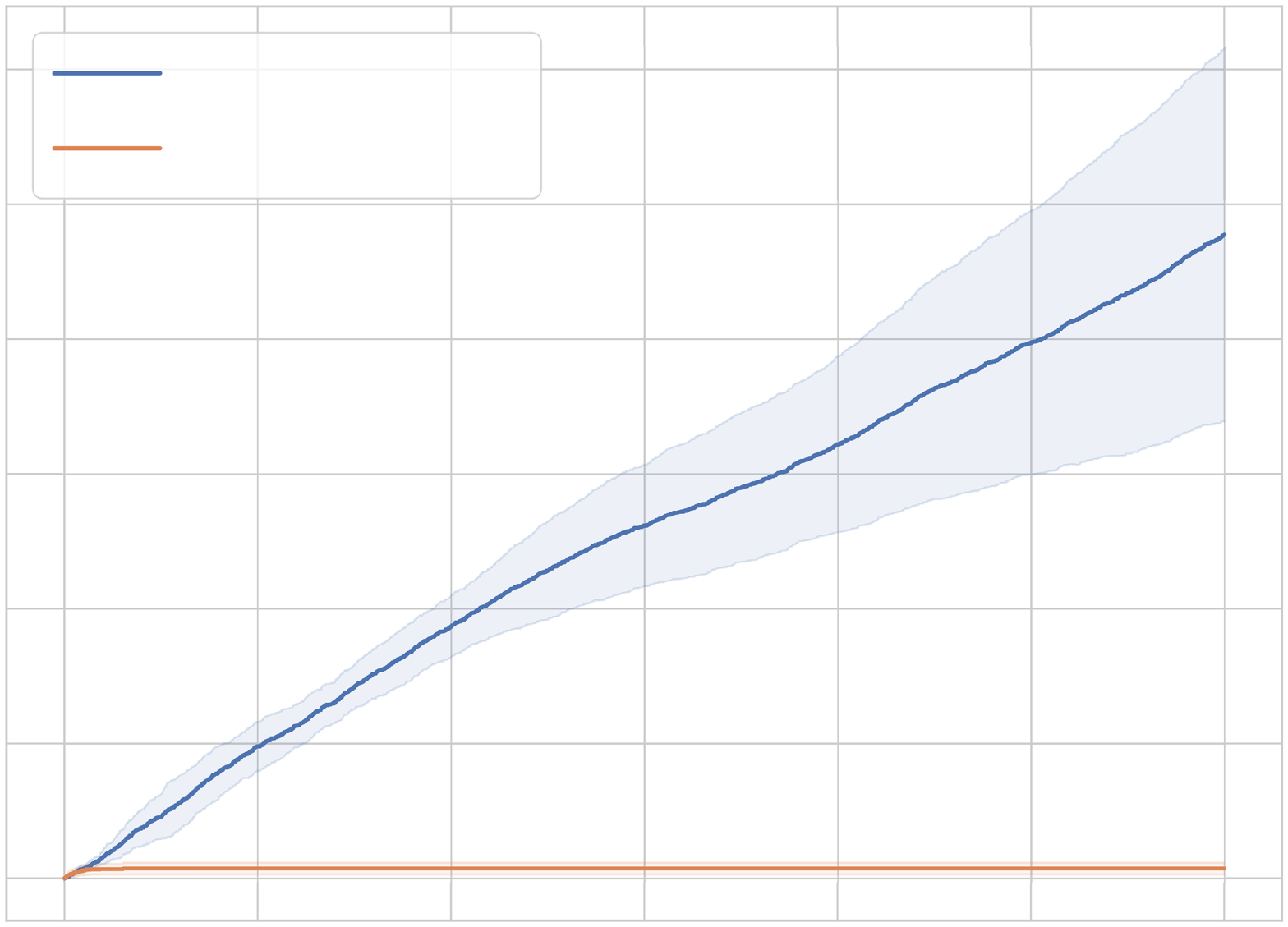}
 	\caption{Sampling comparison between GFlowCausal and CORL.}
	\label{sampling_comparison}
\end{figure}
Next, we show the accumulated number of graphs with high TPR under the same samplings during the training process. We compare GFlowCausal with CORL since they are ordering-based approaches (we apply Case 3 in GFlowCausal). In this setting, we conduct experiments on 30 nodes and 150 edges. We set 5 different seeds and 30 batches with 200 epochs for each sampling. We count the number of graphs with TPR higher than 0.6 and plot the result in Figure~\ref{sampling_comparison}. Since GFlowCausal starts sampling with a uniform policy, there are some fluctuations. In comparison, it is clear that GFlowCausal samples graphs significantly better than CORL. Especially, GFlowCausal generates a few graphs with TPR $> 0.7$ in each seed, while CORL does not obtain such graphs. We guess RL-based exploration tends to be satisfied only with massive enough trajectories. For GFlowCausal, it becomes better with more visiting graphs. These results again validate the advantages of learning a generative policy.

We compare GFlowCausal with other 7 baselines. The result on 30 nodes shows in the Table~\ref{table_1}. We evaluate each algorithm on 5 different graphs and compare the average performance (the values in parenthesis represent the standard deviation across datasets per task). For fairness, we set the same batch size and epochs to keep the total number of graphs consistency. Our approach performs the best in most cases except the SHD in ER5, we conjecture the reason behind is that the pruning techniques 
do not identify some indirect relationships, and thus fully-connected graphs are more likely to have more unnecessary edges. We have also evaluated our method on 100 nodes on ER5 and SF5 in Table~\ref{table_2}. For ER5 case, GraN-DAG, PC and RL-BIC2 could not handle 100 nodes. We can see GFlowCausal performs best in most settings while NOTEARS and CORL are not too far behind. For Non-Gaussian noise, we take the Gumbel distribution as an example experiment to evaluate the performance of our approach (see Appendix~\ref{Linear Models with Non-Gaussian Noises}). For SF5 case, our method achieves the best TPR, and the SHD is not far behind that of CORL. We conjecture the reason behind this is that the pruning output has some randomness. Since both CORL and GFlowCausal output a fully-connected graph first, higher TPR represents our approach generates better ordering and thus grabs the correct causal relationship among variables. Others can not handle such large-scale experiments, and the performances are not satisfying.

\subsection{Nonlinear Models}
\label{nolinearmodels}
In this section, we first consider causal relationship with $f_i$ being a function sampled from a Gaussian Process (GP) with radial basis function kernel of bandwith one. The additive noise follows standard Gaussian distribution, which is known to be identifiable. We use $r_2(s_f)$ to calculate the reward for GFlowCausal. The variable selection used here is the CAM pruning from \cite{buhlmann2014cam}. Since using GP regression to calculate the rewards is time-consuming, we only consider 10 nodes and 40 edges with 1000 samples to generate datasets. 

For comparison, the competitive results in existing works are included as our baselines \cite{zhu2019causal} and are shown in Table~\ref{table_3}. ICA-LiNGAM, NOTEARS, DAG-GNN and PC perform poorly on this causal relationship. GFlowCausal performs the best, and CORL is slightly worse. We conjecture the reason behind this is that both CORL and GFlowCausal are general architecture in which reward functions are not limited. GFlowCausal takes advantage of the efficient ordering search and strong generation ability; thus it is more likely to sample better graphs. We also consider causal relationships with quadratic functions in Appendix~\ref{Nonlinear Models with Quadratic Functions}.

\begin{table*}\small
\centering
\renewcommand\arraystretch{1.8}
\setlength{\abovecaptionskip}{0.1cm}
\setlength{\belowcaptionskip}{-0.1 cm}
\setlength{\tabcolsep}{0.9 mm}{
\caption{Nonlinear results on Gaussian Process data models with 10-node graphs }
\label{table_3}
\begin{tabular}{ccccccccccccc} 
\toprule & &   ICA-LiNGAM     &  NOTEARS          &   DAG-GNN        &  GraN-DAG       & PC &  RL-BIC2          & CORL   & Ours        \\ \hline
&TPR   &   0.63$\pm$0.07    &   0.18$\pm$0.09   &  0.07$\pm$0.03  &  0.81$\pm$0.05  & 0.08$\pm$0.04 &  0.80$\pm$0.09    &   0.91$\pm$0.02     &   \textbf{0.93$\pm$0.01}  \\ 
&SHD   &   48.4$\pm$6.56    &   12.0$\pm$5.18   &  34.6$\pm$1.36  &  10.2$\pm$2.93 & 47.5$\pm$8.23 &  7.8$\pm$4.2  &    4.7$\pm$2.1    &    \textbf{3.4$\pm$2.3}\\
\bottomrule
\end{tabular}}
\end{table*}

\subsection{Experimental Results on Real Dataset}
\label{Experimental Results on Real Dataset}
The Sachs dataset \cite{sachs2005causal}, with 11-node and 17-edge true graph is widely used for research on graphical models. The observational data set has $n = 853$ samples and is used to discover the causal structure. The empty graph has SHD 17. We use Gaussian Process regression to model the causal relationships in calculating the score. In this experiment, RL-BIC2, CORL achieve the best SHD 11. GraN-DAG, ICA-LiNGAM achieve the best SHDs with 13 and 14 respectively. DAG-GNN and NOTEARS have SHD 16 and 19 respectively. GFlowCausal achieves best SHD 9.

\subsection{Ablation Experiments}
In this section, we conduct two different ablation experiments to show GFlowCausal takes advantage of both diverse samplings due to the training objectives of GFlowNets and particular design of allowed actions for efficiency sampling. In subsection~\ref{experiment1}, we compare GFlowCausal with RL-based approach with the same state, action and reward function. In subsection~\ref{experiment2}, we delete the transitive closure $H$ and masked matrix $M$ for allowed action sets, and incorporate the smooth function into the reward function similarly to the traditional score-based approach. 

\subsubsection{Experiment 1: Compare GFlowCausal with other maximum entropy RL methods}
\label{experiment1}
\begin{figure}

	\centering
	\includegraphics[width=2.5in]{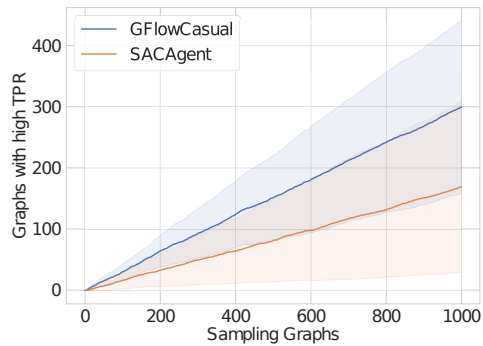}
 	\caption{Ablation comparisons on GFlowCausal and reinforcement learning based approach.}
	\label{sampling_comparison_Ablation}
\end{figure}

To make our proposed method more convincing and demonstrate its powerful exploration ability, we compare GFlowCausal with other maximum entropy RL methods in this section. We run Soft Actor Critic \cite{haarnoja2018soft} with the same setup, where the state, action space, and reward function are the same. We consider the setting of ER5 with 12 nodes. We set 5 different experiments and 10 batches with 3000 epoches for each sampling. The reward function here is \eqref{varsor_r1}. Then based on learned policy we generate new 1000 graphs and count the number of graphs with TPR higher than 0.6. We plot the results of re-sampling in Figure~\ref{sampling_comparison_Ablation}. 

In our experiment, we find the SACAgent could not always attain the convergence. In addition, we notice that after the training procedure, SACAgent could only sample sub-optimal graphs, which means the best graph found in the re-sampling process is not always as good as the best graph found in the training process. In contrast, GFlowCausal could always generate better graphs during the re-sampling process and explore diverse candidates. The TPR of GFlowCausal is $0.91\pm0.02$ while that of SACAgent is $0.69\pm0.03$. 

The performance of SAC is poor but not surprising: since it is consistent with reward-maximization, SAC could find a mode to latch onto quickly and concentrates all of its probability mass on that mode. This causes the no-diversity dilemma we are trying to avoid. This circumstance also appears in the discussion section in \cite{bengio2021flow}.

\subsubsection{Experiment 2: Traditional score-based approach using GFlowNets}
\label{experiment2}

\begin{figure}
\centering

\subcaptionbox{}
{
\includegraphics[width=.8\linewidth]{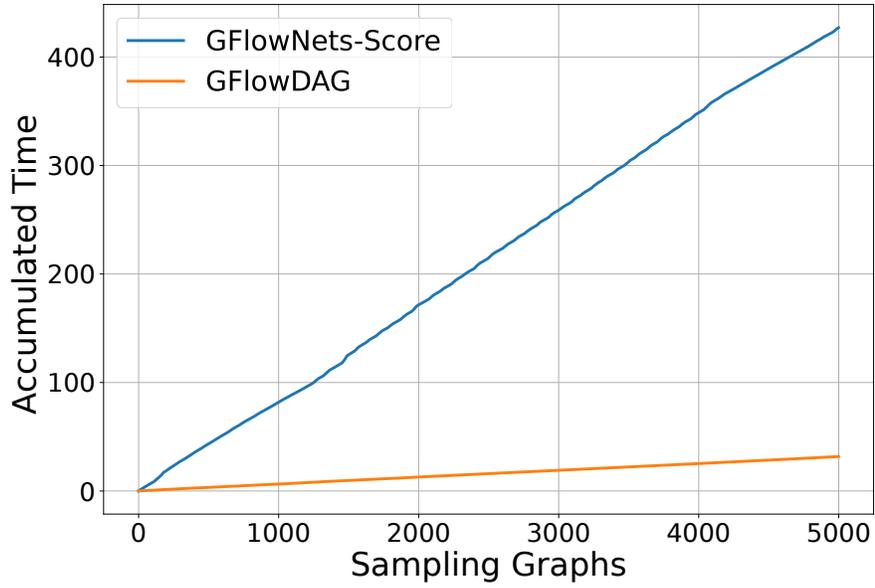}}
\subcaptionbox{}
{
\includegraphics[width=.8\linewidth]{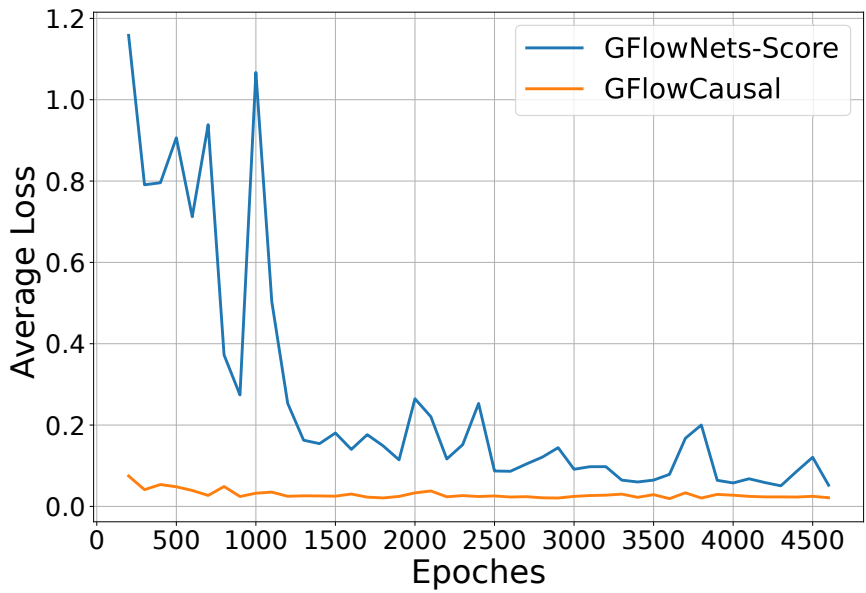}
}

\caption{ (1) Upper: Accumulated time of sampling graphs: GFlowCausal is 10x faster than GFlowNets-Score on 12 nodes setting. (2) Bottom: Average Flow Loss between GFlowNets-Score and GFlowCausal. GFlowNets-Score has more fluctuations and GFlowCausal converges more quickly.   \label{Ablation_2}}

\end{figure}

We also conduct experiment with the traditional score-based approach using GFlowNets, named GFlowNets-Score. In this setting, we applied the reward function incorporating both the score function and acyclicity constraints similarly to \cite{zhu2019causal}. To guarantee the first acyclicity properties in Theorem~\ref{theorem 1}, we also introduce the mask matrix to avoid taking repeated actions or the actions introducing correlation relationships (there are two edges between two nodes). The transitive closure defined in Definition~\ref{transitive_closure} is not used here. The final state is the graph with $\frac{n(n-1)}{2}$ edges since these are the maximum edges in a DAG. Without transitive closure update defined in \eqref{TC_update_1}, we could not guarantee the acyclicity of generated graphs but only give them some penalties. We sample 5000 graphs for ER5 with a 12-node setting in the training process and re-sample 1000 graphs based on the trained policy. We found that GFlowCausal could, on average, sample a graph 10x faster than GFlowNets-Score in Figure~\ref{Ablation_2}. The reason behind this is that the traditional score-based approach uses the smooth function for acyclicity constraints with the exponential complexity that the GFlowCausal could avoid; secondly, by introducing transitive closure $H$ we can further reduce action space. The TPR of GFlowNets-Score is $0.52\pm0.03$, which is also not surprising. We also plot the flow loss in Figure~\ref{Ablation_2} to show GFlowCausal converges faster than GFlowNets-Score since the latter approach has a larger action space and needs to sample a massive number of trajectories.

Above all, we conduct these two ablation experiments to show that we take advantage of the exciting exploration ability of GFlowNets and design a state matrix and action space more suitable for it, thus proposing the powerful GFlowCausal. Finally, we compare the three approaches with the 1000 sampling graphs and show the results in Table~\ref{ablation_combine}. GFlowCausal samples graphs fast, and it could explore diverse high-reward candidates. Since the action space of GFlowNets-Score is much larger than other two approaches, it generates 1000 distinct graphs while the performance is the worst.
\begin{table}[ht]
\centering
\renewcommand\arraystretch{1.4} 
\setlength{\abovecaptionskip}{0.1cm}
\setlength{\belowcaptionskip}{-0.2cm}
\caption{Comparisons among GFlowNets-Score, SACAgent and GFlowCausal with 1000 sampling 12-node graphs.}
\label{ablation_combine}
\begin{tabular}{cccc}
\toprule
 & \makecell[c]{Average\\Sampling time} & \makecell[c]{Distinct\\Graphs} & TPR           \\ \hline
GFlowNets-Score & $0.08\pm0.02$         & \textbf{1000} & $0.52\pm0.03$ \\ \hline
SACAgent        & $0.006\pm0.001$       & 30              & $0.67\pm0.04$ \\ \hline
GFlowCausal     & \textbf{0.005 $\pm$ 0.001}       & 789             & \textbf{0.85 $\pm$ 0.04} \\ 
\bottomrule
\end{tabular}
\end{table}
\subsection{Experiment 3: Compare GFlowCausal with DAG-GFlowNet}
\label{concurrentwork}
There are some overlaps between ours and concurrent work in DAG-GFlowNet~\cite{deleu2022bayesian} (a GFlowNets-based bayesian structure learning method). Both works consider use GFlowNets framework to uncover causal relationships among variables and gradually adding edges to generate DAGs. However, there are some fundamental differences between DAG-GFlowNet and our GFlowCausal, making our approach more efficient and effective in some cases: 


\begin{itemize}
    \item The stopping criteria of DAG-GFlowNet is sampling an invalid action (break the DAG constraints in causal structure learning), thus the potential candidate space is larger than ours. The stopping criteria of GFlowCausal is generating the unique topological sort, and we propose  Theorem~\ref{theorem_2} to help identify these stopping criteria. Therefore, the length of each trajectory in GFlowCausal is shorten than DAG-GFlowNet.
    \item In order to smaller the state and action space, GFlowCausal proposes topological ordering generation approach, which is efficient and outperforms than DAG-GFlowNet.
    \item For every new generated DAG (refers to the state $s_t$), DAG-GFlowNet computes the reward for it. Since the time of reward calculating is dominant, GFlowCausal applies the episodic reward setting, it only calculates the reward for the terminal state ($s_f$). In this way GFlowCausal could save much time on this process.
    \item The reward function in GFlowCausal is either BIC score or varsortability, which are different to DAG-GFlowNet. In our experiments, we find varsortability could become a new benchmark for evaluating causal relationships with some certain assumptions, and it could help us find better graphs. 
    \item Since the sampling output is a fully-connected graph in GFlowCausal, we need to further prune unessential edges. The pruning process could not guarantee the robust results. In contrast, DAG-GFlowNet could generate the exact graph without pruning process.
\end{itemize}

We conduct an experiment on ER5 with 12 nodes and report the results of 3 different seeds in Table~\ref{compare_2}. The batch size is 64 and the number of samples is 1000. Following the setting in~\cite{deleu2022bayesian} we also show the expected SHD of both approaches, which is given by
\begin{equation}
    \mathbb{E}-\text{SHD} \simeq \frac{1}{n}\sum_{k=1}^n \text{SHD}(\mathcal{G}_k,\mathcal{G}^{\star}),
\end{equation}
where $\mathcal{G}^{\star}$ is the ground-truth and $\mathcal{G}_k$ is the $k$-th generated graph. We also compute the \textit{area under the ROC curve} \cite{husmeier2003sensitivity} for comparison and report it as AUROC. The lower $\mathbb{E}$-SHD and higher AUROC, the better the results.
If we set the overall training and sampling time of DAG-GFlowNets as the baseline, our GFlowCausal could save about 70$\%$ of time under 12 nodes. We can expect the time saving is more significant with larger size of dataset.

\begin{table}[ht]
\centering
\renewcommand\arraystretch{1.4} 
\setlength{\abovecaptionskip}{0.1cm}
\setlength{\belowcaptionskip}{-0.2cm}
\caption{Comparisons between GFlowCausal and DAG-GFlowNet}
\label{compare_2}
\begin{tabular}{ccccc}
\hline
                & Overall-time & AUROC &  $\mathbb{E}-\text{SHD}$         \\ \hline
DAG-GFlowNet  &     1x   &  0.6329   &  35.4 \\ \hline
GFlowCausal     &  \textbf{0.3x}      &  \textbf{0.8836}   &  \textbf{31.2} \\ \hline
\end{tabular}
\end{table}

\section {Conclusion}
\label{conclusion}
In this work, we introduce GFlowCausal, a novel and generalized approach for causal discovery, which casts the searching problem into a generation problem. GFlowCausal learns the best policy to generate DAGs with probability proportional to their rewards based on flow networks. Avoiding previous inefficient continuous constrained optimization, we utilize transitive closure to design allowed action, guaranteeing the acyclicity constraint. Theoretical analysis shows its efficiency and consistency between final states and fully-connected graphs, thus reducing the action space. In various experiments, GFlowCausal improves the performance significantly and shows its superiority. The current limitation is that such architecture is only applied in DAG structure, and thus one future direction is to break such constraints and fit into more general situations. Nevertheless, we believe GFlowCausal could provide a reference for future generative model-based approaches.\\
\textbf{Negative Social Impact}\ The authors do not foresee negative social impacts of this work specifically.


\bibliographystyle{unsrt}  
\bibliography{references}  

\newpage

\section{Proof of Main Results}
\subsection{Proof of Theorem~\ref{theorem 1}}
\label{Proof of the Proposition 1}




We first prove for $i,j = 1,...,d$, if $M_{i,j}(s_{t+1}) = 0$, we have $a_{t+1} = v_{j \rightarrow i}^{+} \in \mathcal{A}$, $i,j = 1,...,d$, if $M_{i,j}(s_{t+1}) = 1$, we have $a_{t+1} = v_{j \rightarrow i}^{+} \not\in \mathcal{A}$  by mathematical induction. On the one hand, at state $s_0$, we have $H(s_0) = I_{ d\times d}$ and $A(s_0) = \textbf{0}_{d \times d}$ by definition. Then $M(s_0)= I_{ d\times d}$. It is clear $\forall v_i,v_j$, if $i \neq j $, $v_{i \rightarrow j}^{+}$ does not introduce any cycle, while if $i = j$, there is a self-loop. Hence, 
\begin{align}
  &\text{if}~ M_{i,j}(s_0)=0,~ a_0 = v_{j\rightarrow i}^{+} \in \mathcal{A}, ~\text{and} \nonumber \\
  &\text{if}~ M_{i,j}(s_0)=1,~ a_0 = v_{j\rightarrow i}^{+} \not\in \mathcal{A}.   \nonumber
\end{align}


On the other hand, suppose at state $s_t$ we have  
\begin{align}
\forall M_{i,j}(s_t) &= 0, ~ v_{j\rightarrow i}^{+} \in \mathcal{A}, \text{and} \\
\forall M_{i,j}(s_t) &= 1, ~ v_{j\rightarrow i}^{+} \not\in \mathcal{A}.
\end{align}

If $M_{i,j}(s_{t+1}) = 1$, we have $A_{i,j}(s_{t+1}) = 1$ or $H_{j,i}(s_{t+1}) = 1$ by 
\begin{equation}
    M(s_{t+1})= A(s_{t+1}) \vee H^{\mathsf{T}}(s_{t+1}). \nonumber
\end{equation}
If $ A_{i,j}(s_{t+1}) = 1 $, then $v_{j\rightarrow i} \in \mathcal{E}$. Thus $v_{j\rightarrow i}^{+}$ introduces repeated action, yields $v_{j\rightarrow i}^{+} \not\in \mathcal{A}$. If $H_{j,i}(s_{t+1})=1$, since
\begin{equation}
\label{TC_update}
    H(s_{t+1}) =  H(s_t) \vee h_{j}(s_t)\cdot h^{=}_{i}(s_t),
\end{equation}
then there are two cases: 1) If $H_{j,i}(s_{t})=1$, then $v_i \in \mathcal{V}^{\sharp}(v_j)$; 2) For $\forall a_t = v_{a \rightarrow b}^{+}, a \neq b $, if $ [h_b(s_t) \cdot h_a^{=}(s_t)]_{j,i} = 1$ and $i \neq j$, then based on Lemma~\ref{lemma_3} (proved in Appendix~\ref{Proof of the Lemma 1}) we have $v_{i} \in \mathcal{V}^{\sharp}(v_j)$. Hence we prove $H_{j,i}(s_{t+1}) = 1, j \neq i$, then $v_i \in \mathcal{V}^{\sharp}(v_j)$. 

Similarly, if $M_{i,j}(s_{t+1}) = 0$, we have $A_{i,j}(s_{t+1}) = 0$ and $H_{j,i}(s_{t+1}) = 0$. 
If $A_{i,j}(s_{t+1}) = 0$, then $v_{j \rightarrow i} \not\in \mathcal{E}$. If $H_{j,i}(s_{t+1}) = 0$, then $H_{j,i}(s_t)=0$ and
$\forall a_t = v_{a \rightarrow b}^{+},~ a \neq b ,~ [h_b(s_t) \cdot h_a^{=}(s_t)]_{j,i} = 0, i \neq j$.
Then based on Definition~\ref{transitive_closure} and Lemma~\ref{lemma_3}, we have $v_i\not\in \mathcal{V}^{\sharp}(v_j)$. Then $v_{j \rightarrow i}$ does not introduce any cycle.

\begin{lemma}
\label{lemma_3}
Denote $h_{b,j}, j =1,...,d$ as the $j$-th element of the vector $h_{b}$, and $h^{=}_{a,i}, i =1,...,d$ as the $i$-th element of the vector $h^{=}_{a,i}$, such that
\begin{equation}
\label{vec_cdot}
    h_{b}\cdot h^{=}_{a} = ( h_{b,j}\cdot h_{a,i}^{=} )_{d \times d}.
\end{equation}
Then for $\forall a_t = v_{a \rightarrow b}^{+}, a \neq b $, if $h_{b,j}\cdot h_{a,i}^{=} = 1$, we have $v_{i} \in \mathcal{V}^{\sharp}(v_{j})$, if $h_{b,j}\cdot h_{a,i}^{=} = 0$ and $H_{j,i}(s_t)=0$, we have $v_{i} \not\in \mathcal{V}^{\sharp}(v_{j})$.
\end{lemma}

Then $\forall H_{j,i}(s_{t+1}) =1$, we have $v_{j\rightarrow i}^{+} \not\in \mathcal{A}$ since $v_{i} \in \mathcal{V}^{\sharp}(v_{j})$, and $\forall H_{j,i}(s_{t+1}) = 0$, we have $v_{i} \not\in \mathcal{V}^{\sharp}(v_j)$ based on ~\ref{TC_update} and Lemma~\ref{lemma_3}. Hence, we have $\forall M_{i,j}(s_{t+1}) = 1, ~ v_{j\rightarrow i}^{+} \not\in \mathcal{A}$, and $\forall M_{i,j}(s_{t+1}) = 0,~ v_{j\rightarrow i}^{+} \in \mathcal{A}$.
We can conclude that 
\begin{align}
  &\text{if}~ M_{i,j}(s)=0,~ a = v_{j\rightarrow i}^{+} \in \mathcal{A}, ~\text{and} \nonumber \\
  &\text{if}~ M_{i,j}(s)=1,~ a = v_{j\rightarrow i}^{+} \not\in \mathcal{A}.   \nonumber
\end{align}



Next, we prove two acyclicity properties: 

1) Since if $A_{i,j}(s_t) = 1$, we have $M_{i,j}(s_t) = 1$, then $v_{j\rightarrow i}^{+} \not\in \mathcal{A}$. Thus only if $A_{i,j}(s_t) = 0$ corresponding to $v_{j\rightarrow i} \not\in \mathcal{E}$, we have $v_{j\rightarrow i}^{+} \in \mathcal{A}$. Then $A(s_t) \neq A(s_{t+k}),~ k > 0 $. Thus $\forall s_t,s_{t+k} \in \tau$ with $k>0$, $s_t \neq s_{t+k}$.  Then we prove $\exists \tau_i, \tau_j \in \mathcal{T}, \tau_i \neq \tau_j$, such that $s_f \in \tau_i, s_f \in \tau_j$. This is trivial since for any $\tau_i = (s_0,...,s_{f-2},s_{f-1},s_f)$ where $\forall a_{f-2} = v_{k_1\rightarrow k_2}^+$ and $\forall a_{f-1} = v_{k_3\rightarrow k_4}^+$, we can obtain another $\tau_j  = (s_0,...,s_{f-2},\bar s_{f-1},s_f)$ by letting $\forall a_{f-1} = v_{k_1\rightarrow k_2}^+$ and $\forall a_{f-2} = v_{k_3\rightarrow k_4}^+$.





2) Since $\forall H_{i,j} = 1, i \neq j $, we have $ v_{j} \in \mathcal{V}^{\sharp}(v_i)$ and $M_{j,i} = 1$, thus $v_{i\rightarrow j }^{+} \not\in \mathcal{A}$, then we can guarantee if $ v_{j} \in \mathcal{V}^{\sharp}(v_i)$, then $v_i \not\in \mathcal{V}^{\sharp}(v_j) $. 

Then we complete the proof.

\subsection{Proof of Lemma~\ref{lemma_3}}
\label{Proof of the Lemma 3}
We first prove $\forall j \neq b$, $h_{b,j} = 1$ indicates $v_{j} \in \mathcal{V}^{\flat}(v_b)$, and $\forall i \neq a,~ h^{=}_{a,i}=1 $ indicates $v_{i} \in \mathcal{V}^{\sharp}(v_a)$. Since $h_b$ corresponds to the $b$-th column of $H(s_t)$, $h_{b,j}$ corresponds to the $j$-row of $h_{b}$, then $h_{b,j}$ is equivalent to $H_{j,b}$. 
Based on Definition~\ref{transitive_closure}, if $H_{j,b} = 1$ and $b \neq j$, then $v_{b} \in \mathcal{V}^{\sharp}(v_{j})$. Thus $ \forall b \neq j$, $h_{b,j} = 1$ indicates $v_{j} \in \mathcal{V}^{\flat}(v_{b})$. Similarly, we can prove $h^{=}_{a,i}$ corresponds to $H_{a,i}$, thus if $ a \neq i$, $h^{=}_{a,i}=1 $ indicates $v_{i} \in \mathcal{V}^{\sharp}(a)$.

Next we prove $\forall j \neq b$, $h_{b,j} = 0$ indicates $v_{j} \not\in \mathcal{V}^{\flat}(v_b)$, and $i \neq a$, $h_{a,i}^{=} = 0$ indicates $v_i \not\in \mathcal{V}^{\sharp}(v_a)$ by contradiction. Suppose $h^{=}_{a,i}=0$ and $v_i \in \mathcal{V}^{\sharp}(v_a)$. If $v_i \in \mathcal{V}^{\sharp}(v_a)$, we have $H_{a,i} = 1$ based on Definition~\ref{transitive_closure}. Since $h_{a,i}^{=}$ corresponds to $H_{a,i}$, thus $h_{a,i}^{=}=0$ leads to a contradiction. Suppose $h_{b,j} = 0$ and $v_j \in \mathcal{V}^{\flat}(v_b)$. If $v_{j} \in \mathcal{V}^{\flat}(v_b)$, we have $v_b \in\mathcal{V}^{\sharp}(v_j)$ based on Definition~\ref{transitive_closure} , then $H_{j,b} = 1$. Since $h_{b,j}$ is equivalent to $H_{j,b}$, thus $h_{b,j} = 0$ leads to a contradiction.

Then we prove when $a_t = v_{a\rightarrow b}^{+} \in \mathcal{A}$, if $h_{b,j}\cdot h_{a,i}^{=} = 1$, then $v_{i} \in \mathcal{V}^{\sharp}(v_{j})$. Considering $h_{b,j}\cdot h_{a,i}^{=} = 1$, we have four following cases:
\begin{itemize}
    \item If $j = b, i = a$, 
    then $a_t = v_{a \rightarrow b}^{+} = v_{i \rightarrow j}^{+}$. Hence $v_{i} \in \mathcal{V}^{\sharp}(v_{j})$.
    \item If $j = b, i \neq a$, then $v_{j} = v_b$, $v_{i} \in \mathcal{V}^{\sharp}(v_a)$. Then $v_{i} \in \mathcal{V}^{\sharp}(v_{j})$ since $v_{a} \in \mathcal{V}^{\sharp}(v_{b})$.
    \item If $j \neq b, i = a$, then $v_{j} \in \mathcal{V}^{\flat}(v_b)$ and $v_{i} = v_a$,  thus $v_{i} \in \mathcal{V}^{\sharp}(v_{j})$.
    \item If $j \neq b, i \neq a$, then $v_{i} \in \mathcal{V}^\sharp(v_a), v_{j} \in \mathcal{V}^\flat(v_b) $, thus $v_{i} \in \mathcal{V}^{\sharp}(v_{j})$.
\end{itemize}

Next we prove if $h_{b,j}\cdot h_{a,i}^{=} = 0$, and  $H_{j,i}(s_t)=0$, then $v_{i} \not\in \mathcal{V}^{\sharp}(v_j)$ .
We should note if $j = b, i = a$,  then $h_{b,j} = 1$ and $h^{=}_{a,i} = 1$, which contracts to the statement. Thus if $h_{b,j}\cdot h_{a,i}^{=} = 0$, then $j = b $ and $i = a$ can not occur simultaneously. Since $H_{j,i}(s_t)=0$, then $v_i \not\in \mathcal{V}^\sharp(v_j)$ at $s_t$. Then we should check whether $a_t = v_{a\rightarrow b}^{+}$ could connect $v_i$ and $v_j$.
\begin{itemize}
    \item If $j = b, i \neq a$, then $v_j = v_b , h_{a,i}^{=} = 0 $. Thus $v_i \not\in \mathcal{V}^{\sharp}(v_j)$ since $v_i \not\in \mathcal{V}^{\sharp}(v_a)$.
    \item If $j \neq b, i = a$, then $h_{b,j}= 0$, $v_i = v_a$. Thus $v_i \not\in \mathcal{V}^{\sharp}(v_j)$ since $v_j \not\in \mathcal{V}^{\flat}(v_b)$.
    \item If $j \neq b$, $i \neq a$ and $h_{b,j}=0, h_{a,i}^{=} \neq 0$, we have $v_j \not\in \mathcal{V}^{\flat}(v_b), v_i \in \mathcal{V}^{\sharp}(v_a)$, then $v_{a\rightarrow b}^{+}$ does not connect $v_i$ and $v_j$.
    \item If $j \neq b$, $i \neq a$ and $h_{b,j}\neq 0, h_{a,i}^{=} = 0$, we have $v_j \in \mathcal{V}^{\flat}(v_b), v_i \not\in \mathcal{V}^{\sharp}(v_a)$, then $v_{a\rightarrow b}^{+}$ does not connect $v_i$ and $v_j$.
    \item If $j \neq b$, $i \neq a$ and $h_{b,j}=0, h_{a,i}^{=} = 0$, then $v_{a\rightarrow b}^{+}$ does not connect $v_i$ and $v_j$ since $v_i \not\in \mathcal{V}^{\sharp}(v_a)$ and $v_j \not\in \mathcal{V}^{\flat}(v_b)$.
\end{itemize}

Then we complete the proof.

\subsection{Proof of Theorem~\ref{theorem_2}}
\label{proof of the Proposition 2}

We first prove if $\vec{\textbf{v}}(s_t) = {\vec{\textbf{v}}_{\star}}$, then $\forall i,j = 1,...,d, ~ Q_{i,j}(s_t) \neq 0$. Suppose $\vec{\textbf{v}}(s_t) = {\vec{\textbf{v}}_{\star}}$, then $\forall i,j = 1,...,d$, if $v_j \in \mathcal{V}^{\sharp}(v_i)$, then $H_{i,j} (s_t) = 1$. Hence $Q_{i,j}(s_t) \neq 0$ based on \eqref{eq_11}; if $v_i \in \mathcal{V}^{\sharp}(v_j)$ we have $H_{j,i} (s_t) = 1$, thus $Q_{i,j}(s_t) \neq 0$ since $H_{i,j}^{\mathsf{T}} (s_t) = 1$; if $v_i = v_j$, we obviously have $Q_{i,j}(s_t) \neq 0$.

Then we prove  if $\forall i,j= 1,...,d, ~ Q_{i,j}(s_t) \neq 0$, then $\vec{\textbf{v}}(s_t) = {\vec{\textbf{v}}_{\star}}$ by contradiction. If $\exists Q_{i,j}(s_t) = 0 $, then $H_{i,j}(s_t) =0$ and $H_{j,i}(s_t) = 0$, then $v_i \not\in \mathcal{V}^{\sharp}(v_j) , v_j \not\in \mathcal{V}^{\sharp}(v_i)$. Then we can see $v_i$ and $v_j$ are not in $\vec{\textbf{v}}(s_t)$ simultaneously. Thus $\vec{\textbf{v}}(s_t) \neq \vec{\textbf{v}}_{\star}$.

Above all, we complete the proof for $\vec{\textbf{v}}(s_t)$ is identified iff $\forall i,j= 1,...,d, Q_{i,j} \neq 0$.




Second, we prove once $\vec{\textbf{v}}(s_{t})$ is identified, then $\forall k > 0 , \vec{\textbf{v}}(s_{t+k}) =\vec{\textbf{v}}(s_{t}) =  \vec{\textbf{v}}_\star $ by contradiction. Suppose $\vec{\textbf{v}}(s_{t})  = {\vec{\textbf{v}}_{\star}},~ \forall a_{t} = v_{k_i \rightarrow k_j}^{+} \in \mathcal{A}$. Then, $\vec{\textbf{v}}(s_{t+1})$ contains $m$ elements, $m \ge d$.
1) If $m > d$, then $\vec{\textbf{v}}(s_{t+1})$ must have a cycle, which conflicts with $a_{t} \in \mathcal{A}$;
2) If $m = d$, if $\vec{\textbf{v}}(s_{t+1}) \neq {\vec{\textbf{v}}_{\star}}$, then in ${\vec{\textbf{v}}_{\star}}, \exists v_a < v_b$ such that in $\vec{\textbf{v}}(s_{t+1}), v_b < v_a$. Hence $a_{t} = v_{k_i \rightarrow k_j}^{+} \not\in \mathcal{A}$ since it introduces a cycle.  In summary, $\vec{\textbf{v}}(s_{t+1}) \neq {\vec{\textbf{v}}_{\star}}$ does not hold. 
Further, by recursion we have $\vec{\textbf{v}}(s_{t+k}) =\vec{\textbf{v}}(s_{t}) =  \vec{\textbf{v}}_\star$.


Third, we prove once $\vec{\textbf{v}}(s_t)$ is identified, then $H(s_t) - I_{d \times d}$ leads to a fully-connected graph.  This proof is trivial. Suppose $\vec{\textbf{v}}(s_t) = (v_{k_1},...,v_{k_d} )$ is identified, then $\forall i \neq k_1$, we have $\|h_{v_{k_1},i}\|_0 = d-1$, while $\forall i \neq k_1$ and $ i \neq k_2$, we have $\|h_{v_{k_2},i}\|_0 = d-2$. Hence, by recursion $H(s_t) - I_{d \times d}$ has $d(d-1)/2$ positions of value 1, which is the same as the number of all possible edges. We can easily obtain it leads to a fully connected graph. Finally, we complete the proof.

\subsection{Proof of Lemma~\ref{lemma_1}}
\label{Proof of the Lemma 1}

We prove this lemma by contradiction.
Suppose we can identify $\vec{\textbf{v}}_{\star}$ with less than $(d-1)$ edges. Since $\vec{\textbf{v}}_{\star}$ contains $d$ nodes and it is clear that at least $(d-1)$ edges could connect $d$ nodes, which contradicts to the statement.

Next, we suppose more than $\frac{d(d-1)}{2}$ could identify identify the $\vec{\textbf{v}}_{\star}$. Since for a fully-connected graph there are $\frac{d(d-1)}{2}$ edges, then $\forall i \neq j, $ if $A_{i,j} = 0$ , then $A_{j,i} = 1$. Thus $\forall a_f = v_{i \rightarrow j}^{+}$ will introduce a cycle, which contradicts to the statement. Then we complete the proof.


\subsection{Complexity Analysis}
\label{Proof of the Lower Bound}

Suppose that we need $(d-1+k)$ transitions to identify the topological sort, where $0\leq k \leq \frac{(d-2)(d-1)}{2}$, then there are $C^k_{K}$ different combinations, where $K = \frac{(d-2)(d-1)}{2}$.  Thus we have total $ \sum_{k = 0}^{K} C^k_{K}$ cases. The probability of each transition can be considered as $P(k) = \frac{C^k_{K}}{\sum_{k = 0}^{ K} C^k_{K}}$. Without loss of generality, we assume $K$ is an even number. Then, the expected transition number to identify topological sort is given by
\begin{align}
&~~~~  \sum_{k=0}^{K} P(k) \times (d-1+k)  =\sum_{k=0}^{K}  \frac{C^k_{K}}{\sum_{k = 0}^{ K} C^k_{K}} \times (d-1+k)\nonumber  \\
    & = \frac{1}{2^K}\times \sum_{k=0}^{K} [ C^k_{K}\times (d-1+k) ] = d-1 + \frac{1}{2^K} \sum_{k=0}^{K} k  C^k_{K}  \nonumber\\
    & = d-1 + \frac{1}{2^K}  \left\{  K \times 2^{K-1} +  K + \frac{2K -K^2}{4}C^{\frac{K}{2}}_{K}  \right\}  \nonumber\\
    & \lessapprox \frac{(d+2)(d-1)}{4}. \nonumber 
\end{align}
Hence, the overall complexity for a trajectory is $\mathcal{O}(d^2/4)$.

\section{Baselines}
\label{Baselines}
\begin{itemize}
    \item ICA-LiNGAM \cite{shimizu2006linear} assumes linear Non-Gaussian additive model for data generating procedure and applies independent component analysis to recover the weighted adjacency matrix.  Codes are available at \href{https://sites.google.com/site/sshimizu06/lingam.}{https://sites.google.com/site/sshimizu06/lingam}
    
    \item S \cite{shimizu2006linear} recovers the causal graph by estimating the weighted adjacency matrix with the least squares loss and the smooth characterization for acyclicity constraint, followed by thresholding on the estimated weights. Codes are available at \href{https://github.com/xunzheng/notears. }{https://github.com/xunzheng/notears}
    \item DAG-GNN \cite{yu2019dag} formulates causal discovery in the framework of variational autoencoder. It uses a modified smooth characterization for acyclicity and optimizes a weighted adjacency matrix with the evidence lower bound as loss function. Codes are available at \href{https://github.com/fishmoon1234/DAG-GNN }{https://github.com/fishmoon1234/DAG-GNN}
    \item PC \cite{kalisch2007estimating} is a classic causal discovery algorithm based on conditional independence tests. Implementations of the method are available through the py-causal package at \href{https://github.com/bd2kccd/py-causal}{https://github.com/bd2kccd/py-causal} , written in highly optimized Java codes.
    \item GraN-DAG \cite{lachapelle2019gradient} models the conditional distribution of each variable given its parents with feed-forward NNs. It also uses the smooth acyclicity constraint from NOTEARS. Codes are available at \href{https://github.com/kurowasan/GraN-DAG}{https://github.com/kurowasan/GraN-DAG}
    \item RL-BIC2 \cite{zhu2019causal} formulates the causal discovery as a one-steo decision making process, and combines the score function and acyclicity constraint from NOTEARS as the reward for a direct graph. Codes are available at \href{https://github.com/huawei-noah/trustworthyAI}{https://github.com/huawei-noah/trustworthyAI}
    \item CORL\cite{wang2021ordering} formulates the causal discovery as causal ordering search problem. Codes are available at \href{https://github.com/huawei-noah/trustworthyAI}{https://github.com/huawei-noah/trustworthyAI}
    \item DAG-GFlowNet \cite{deleu2022bayesian} formulates the bayesian structure learning with GFlowNets, in which the edges are added gradually to construct direct acyclic graphs. Codes are available at \href{https://github.com/tristandeleu/jax-dag-gflownet}{https://github.com/tristandeleu/jax-dag-gflownet}.
\end{itemize}

\section{Additional Results}
\label{Additional Results}

\begin{table*}[!ht]\small
\centering
\renewcommand\arraystretch{1.4} 
\setlength{\abovecaptionskip}{0.1cm}
\setlength{\belowcaptionskip}{-0.2cm}
\setlength{\tabcolsep}{0.6 mm}{
\caption{Empirical results on LG data models with 100-node graphs }
\label{table_4}
\begin{tabular}{cccccccccccc} 
\hline & &   ICA-LiNGAM   &  NOTEARS &  DAG-GNN & GraN-DAG  & PC &   RL-BIC2 & CORL  & Ours        \\ \hline
\multirow{2}{*}{SF5}&TPR&  0.64$\pm$0.03  &  0.83$\pm$0.16  &  0.81$\pm$0.14  &  0.06$\pm$0.02 & 0.11$\pm$0.04 &   0.12$\pm$0.02  & 0.94$\pm$0.04  & \textbf{0.96$\pm$0.02}  \\

&SHD&  707$\pm$43    &  200$\pm$41   &  172$\pm$47   & 557$\pm$34   &  547$\pm$26  &   411$\pm$64 &  \textbf{47$\pm$17} & 58$\pm$19 \\
\bottomrule
\end{tabular}}
\end{table*}

\subsection{Linear Models with Non-Gaussian Noises}
\label{Linear Models with Non-Gaussian Noises}
We take the Gumbel distribution as an example experiment to evaluate the performance of our approach. Table~\ref{table_5} shows the results of the Linear model with Gumbel distribution. In this setting we set 12 nodes and 24 edges. Since PC does not perform well in the previous experiments, we do not consider it in this setting. The reward function here is $r_{\text{var}}(s_f,X)$ defined in \eqref{varsor_r1} and we also apply it into all RL-based approaches for fairness. In this experiment, GraN-DAG is not suitable for this relationship. Our approach is superior to all algorithms.

\begin{table*}[!ht]\small
\centering
\renewcommand\arraystretch{1.4} 
\setlength{\abovecaptionskip}{0.1cm}
\setlength{\belowcaptionskip}{-0.2cm}
\setlength{\tabcolsep}{1.65 mm}{
\caption{Empirical results on Linear-Gumbel models with 12-node graphs }
\label{table_5}
\begin{tabular}{cccccccccccc} 
\hline & &   ICA-LiNGAM     &  NOTEARS          &   DAG-GNN        &  GraN-DAG        &  RL-BIC2          & CORL   & Ours        \\ \hline
\multirow{2}{*}{ER2}&TPR   &  0.77$\pm$0.18   & 0.88$\pm$0.09    &  0.82$\pm$0.1  &   0.08$\pm$0.04  &    0.91$\pm$0.03  &  0.94$\pm$0.05     &  \textbf{1$\pm$0}   \\ 
&SHD   &   13.0$\pm$5.5  &  3.4$\pm$6.2   &   7.5$\pm$2.4 &    25.2$\pm$4.8      &  5.8$\pm$4.3   &    4.5$\pm$3.2    &  \textbf{0$\pm$0}  \\
\bottomrule
\end{tabular}}
\end{table*}

\subsection{Nonlinear Models with Quadratic Functions}
\label{Nonlinear Models with Quadratic Functions}
We also conduct Quadratic models to corroborate the generalization of GFlowCausal.  We report the results with 10-node graphs with 40 edges in Table~\ref{table_6}, which shows that our method achieves the best performance. RL-based methods could also achieve good enough performance due to their strong searchability while others do not perform well.
\begin{table*}[!ht]\small
\centering
\renewcommand\arraystretch{1.4} 
\setlength{\abovecaptionskip}{0.1cm}
\setlength{\belowcaptionskip}{-0.2cm}
\setlength{\tabcolsep}{1.6 mm}{
\caption{Empirical results on quadratic models with 10-node graphs }
\label{table_6}
\begin{tabular}{cccccccccccc} 
\hline & &   ICA-LiNGAM     &  NOTEARS          &   DAG-GNN        &  GraN-DAG        &  RL-BIC2          & CORL   & Ours        \\ \hline
&TPR   & 0.76$\pm$0.09    &   0.70$\pm$0.15   & 0.55$\pm$0.14   &  0.73$\pm$0.16   &   0.98$\pm$0.04   &   0.98$\pm$0.02     & \textbf{0.99$\pm$0.01}   \\ 
&SHD   &   20.4$\pm$5.0  &   14.8$\pm$3.4  &   18.0$\pm$2.5 &  39.6$\pm$5.9  &  0.6$\pm$1.2   &     0.6$\pm$0.4     &   \textbf{0.5$\pm$0.5} \\
\bottomrule
\end{tabular}}
\end{table*}


\section{Experiment Environment}
\begin{table*}[!ht]
\renewcommand\arraystretch{1.4} 
\caption{Hyperparameters in experiments.}
  \label{para-table-1}
  \centering
  \begin{tabular}{cccccccccc}
    \toprule
    {Hyperparameters}  & GFlowCausal-30 & GFlowCausal-100 &  CORL-30 & CORL-100  \\ 
    \midrule
    Minibatch size     & 64 & 128 & 64 & 128  \\
    Iterations    & 20000  & 50000 & 20000  & 50000   \\
    Learning rate   & $1e^{-4}$ & $5e^{-4}$ & - &  - \\
    Actor learning rate & - & - & $1e^{-4}$ &  $1e^{-4}$ \\
    Critic learning rate & - & - & $1e^{-3}$ &  $1e^{-3}$ \\
    Model   & MLP & CNN/VGG-16 & MLP+LSTM & Transformer+LSTM\\
    $\#$ embed, hidden units & \multicolumn{4}{c}{256} \\
    Encoder heads   & - & - & - & 8 \\
    Encoder blocks   & - & - & - & 3\\
    Optimizer, $\beta$          & \multicolumn{4}{c}{Adam ( 0.9, 0.999 )}\\
    Data samples    & \multicolumn{4}{c}{1000}   \\
    \bottomrule
  \end{tabular}
\end{table*}
All experiments were conducted on a NVIDIA Quadro RTX 6000 environment with Pytorch. The parameters of GFlowCausal and CORL are shown in Table~\ref{para-table-1}. For other approaches, we set the parameters consistent with the best setting in the original paper. 
For GFlowCausal, a simple 2-MLP with hidden units 256 is enough for 30 nodes. For 100-nodes graphs, we use VGG-16 similar to \cite{simonyan2014very} here.
For CORL, the MLP encoder module consists of 2-layer feed-forward neural
networks with 256 units. The Transformer consists of a feed-forward layer with 256 units and three blocks. Each block
is composed of a multi-head attention network with 8 heads
and 2-layer feed-forward neural networks with 1024 and 256
units, and each feed-forward layer is followed by a normalization layer. The LSTM takes a state as input and outputs an embedding with hidden units 256.

\end{document}